\documentclass{article}

\PassOptionsToPackage{numbers, sort&compress}{natbib}

\usepackage[main, final]{neurips_2026}

\usepackage[utf8]{inputenc} 
\usepackage[T1]{fontenc}    
\usepackage{hyperref}       
\usepackage{url}            
\usepackage{booktabs}       
\usepackage{amsfonts}       
\usepackage{nicefrac}       
\usepackage{microtype}      
\usepackage{xcolor}         

\usepackage{amsmath}
\usepackage{amssymb}
\usepackage{mathtools}
\usepackage{amsthm}

\usepackage{cleveref}

\crefname{proposition}{Prop.}{Props.}
\crefname{section}{Sec.}{Sec.}
\crefname{appendix}{App.}{App.}
\crefname{table}{Tab.}{Tabs.}
\crefname{figure}{Fig.}{Figs.}
\crefname{equation}{Eq.}{Eqs.}

\usepackage{pifont}
\newcommand{\cmark}{\ding{51}}
\newcommand{\xmark}{\ding{55}}

\newcommand{\doublelinecell}[2][c]{\begin{tabular}[#1]{@{}c@{}}#2\end{tabular}}

\theoremstyle{plain}
\newtheorem{theorem}{Theorem}[section]
\newtheorem{proposition}[theorem]{Proposition}

\theoremstyle{definition}

\theoremstyle{remark}

\newcommand{\blue}[1]{{\color{black}#1}}

\newcommand{\inlineeq}[1]{%
  \refstepcounter{equation}%
  \ensuremath{#1}\hfill(\theequation)%
}

\usepackage{graphicx}   
\usepackage{subcaption} 
\usepackage{placeins}

\title{Positional Encoding is All You Need for Scalable Equivariance Constraint Relaxation}

\usepackage[noblocks]{authblk}

\author[1]{Hagay Michaeli}
\author[1,2]{Haggai Maron}
\author[1]{Daniel Soudry}

\affil[1]{Technion -- Israel Institute of Technology}
\affil[2]{NVIDIA}

\begin{document}

\maketitle

\begin{abstract}
Equivariant neural networks incorporate known symmetries of the data, such as permutations, rotations, or translations, directly into their architecture, often improving sample efficiency and generalization. 
However, strict equivariance can make optimization difficult, for example, when the data only approximately satisfy the assumed symmetry, or when the equivariance constraints induce a challenging loss landscape. 
Recent studies have shown that relaxing equivariance during training by introducing non-equivariant dense linear layers can ease optimization in such cases and improve the performance of the resulting strictly equivariant model, after removing the relaxation layers at test time.
While effective, this approach incurs a significant computational overhead, limiting its applicability in scalable architectures. 
In this work, we propose a more efficient constraint-relaxation method based on positional encoding (PE). 
Our framework injects a positional encoding with learnable scale that breaks the model’s original symmetry.
This positional signal is gradually annealed to zero, recovering the original equivariant model by the end of training.
Empirically, we demonstrate significant improvements for translational and rotational equivariant models with only marginal compute and memory overhead.
\end{abstract}

\section{Introduction}
\vskip -0.05 in

\begin{figure}[ht]
  \centering
  \includegraphics[width=\linewidth]{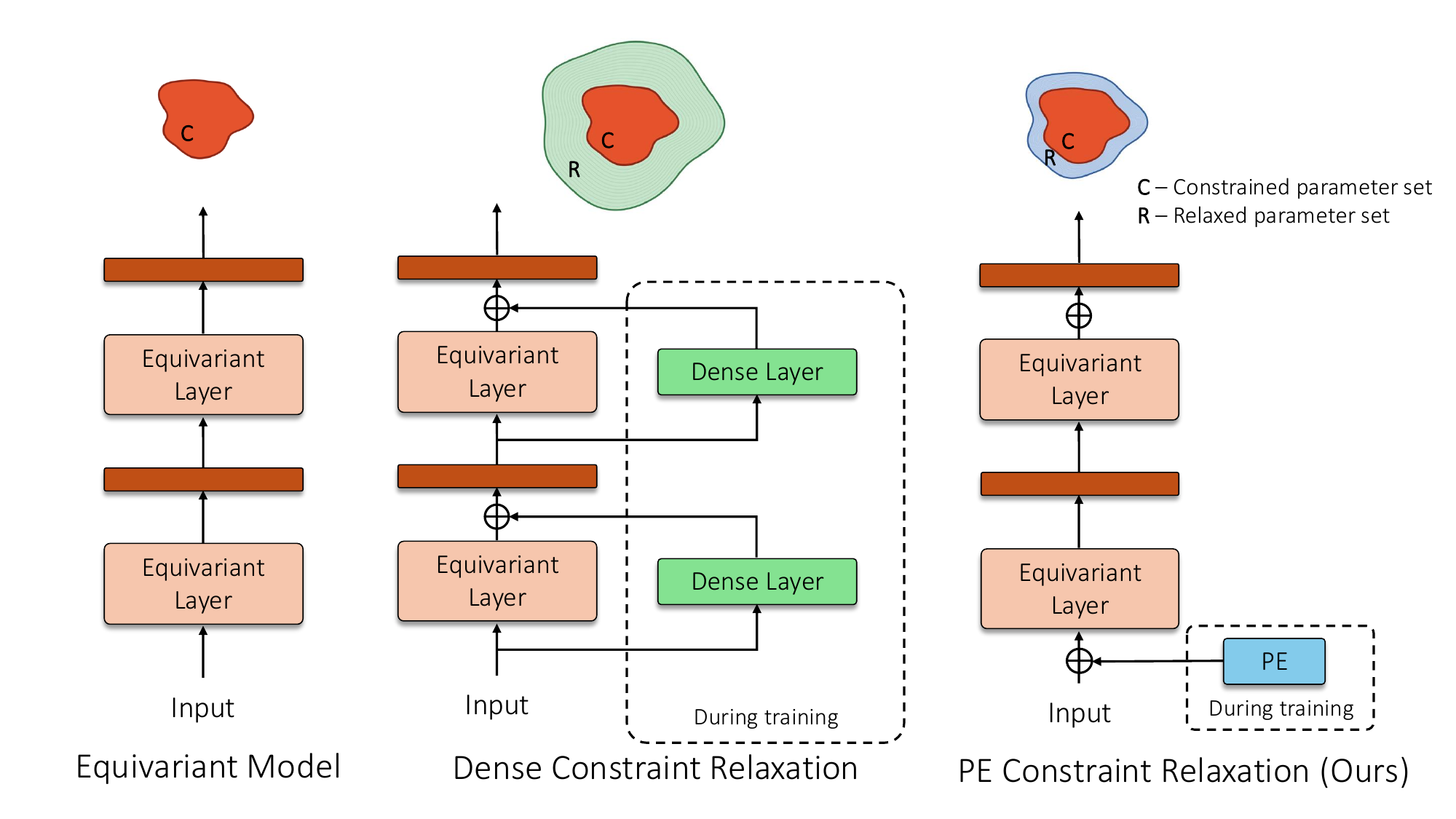}
  \caption{
\textbf{Equivariance constraint relaxation during training.}
\emph{Left:} A strictly equivariant model, whose hypothesis class is restricted to equivariant functions.
\emph{Middle:} Dense constraint relaxation \citep{pertigkiozoglou_improving_2025,pertigkiozoglou_recurrent_2026,manolache_learning_2025} adds a non-equivariant dense branch in parallel to every equivariant layer during training.
This greatly enlarges the hypothesis class, but substantially increases the parameter count and computational cost of training.
\emph{Right:} Our PE constraint relaxation instead adds a single positional encoding to the input.
This relaxes the constraint with negligible parameter and compute overhead.
In both relaxation methods, the auxiliary component is used only during training and is annealed to zero, recovering the original strictly equivariant architecture at inference.
  }
  \label{fig:method}
  \vskip -0.15 in

\end{figure}

Many learning tasks involve symmetries, i.e., transformations of the input that induce predictable transformations of the output. 
Examples include translations in images, rotations in 3D geometry, and permutations in sets and graphs. 
Most equivariant and invariant neural networks enforce this structure by restricting layers to commute with a group action, 
often improving sample efficiency and generalization.
Classic instances include group-equivariant convolutions for images \cite{cohen_group_2016}, steerable CNNs for continuous rotation groups \cite{weiler_learning_2018}, and permutation-equivariant models for sets and graphs \cite{zaheer_deep_2017, maron_invariant_2018},
useful in applications such as 3D perception and physics-inspired learning \cite{Qi2016, satorras_en_2021, deng_vector_2021}.

Despite their advantages, strictly equivariant architectures can be difficult to optimize.
Even when an equivariant optimum exists, structured equivariant parameterizations can impede the loss landscape geometry, causing training to stall at stationary points that unconstrained models can escape \cite{xie_tale_2025,nordenfors_optimization_2024}.
When the data only approximately satisfy the assumed symmetry, strict equivariance induces an approximation-generalization trade-off: the same inductive bias that improves generalization under matched symmetries can become restrictive under symmetry misspecification \cite{wang_approximately_2022, pacini_universality_2025}.
Motivated by these barriers, a growing literature studies approximate or relaxed equivariance \cite{ouderaa_relaxing_2022,romero_learning_2023, huang_approximately_2023, berndt_approximate_2026, elhag_relaxed_2026}.
Recent studies have improved trainability by relaxing equivariance early in training and progressively re-imposing the constraint to recover a strictly equivariant model at inference \cite{pertigkiozoglou_improving_2025,manolache_learning_2025, pertigkiozoglou_recurrent_2026}.
These methods add unconstrained dense branches in parallel to equivariant layers and anneal them to zero, which increases parameter count and training cost and can be prohibitive at scale.

In this work, we propose a simple, scalable alternative: train-time constraint relaxation via positional encoding (PE; \Cref{fig:method}).
Positional encodings are a standard tool for injecting order or geometry into symmetry-preserving architectures, such as Transformers \cite{vaswani_attention_2017, dosovitskiy_image_2020} and graph neural networks (GNNs) \cite{wang_equivariant_2022, dwivedi_graph_2022}.
They are employed to increase the expressivity of such architectures \cite{yun_are_2020, abboud_surprising_2021, sato_random_2021} and, more recently, to control the symmetry of equivariant models \cite{goel_any-subgroup_2025, lawrence_improving_2025, ashman_approximately_2024}.
Building on this direction, we use PE as a transient optimization aid that is annealed to zero, recovering the original strictly equivariant model at inference.
We evaluate our approach on five architectures: Vector Neurons \cite{deng_vector_2021} and Rapidash \cite{vadgama_probing_2025} for rotation-equivariant point-cloud classification, Alias-Free XCiT \cite{michaeli_alias-free_2025} for shift-invariant image classification, E2CNN \cite{weiler_general_2019} for rotation-equivariant smoke plume forecasting \cite{wang_approximately_2022}, and SEGNN \cite{brandstetter_geometric_2022} for E(3)-equivariant N-body forecasting.
Across these models, PE relaxation matches or improves the strict baselines while preserving their equivariance at inference.

\vskip -0.05 in
\section{Related work}
\vskip -0.05 in
Symmetry-aware architectures are a central way to improve sample efficiency and generalization.
A foundational direction involves strictly equivariant layers, for example, group convolutions for grid-structured data \cite{cohen_group_2016} and steerable convolutions for continuous groups \cite{weiler_learning_2018}.
These principles were subsequently expanded to handle other symmetries, including permutation-invariant models for sets and graphs \cite{Qi2016, zaheer_deep_2017, maron_invariant_2018}, $E\left( n \right)$-equivariant GNNs \cite{satorras_en_2021}, and $\mathrm{SO}\left( 3 \right)$ equivariant models for 3D data \cite{deng_vector_2021}.
Recent studies have also strengthened the theoretical understanding of equivariant models, including their expressive power, universality, and generalization properties, while extending exact-equivariant design to broader architectures \cite{dym_universality_2020,finkelshtein_simple_2022, pacini_separation_2024, pacini_universality_2025, islam_platonic_2025, elesedy_provably_2021}.
Beyond generalization, such models can also significantly improve parameter efficiency, for example in deep weight-space models \cite{navon_equivariant_2023}.

\textbf{Approximate equivariance and constraint relaxation.}
Despite their advantages, equivariant models can impose overly strict inductive biases that complicate training.
First, strictly enforcing symmetry can be detrimental when the underlying problem only approximately satisfies the assumed symmetry \cite{wang_approximately_2022}, or when the data distribution itself exhibits symmetry bias \cite{lawrence_augment_2026}.
Additionally, recent studies show that strictly equivariant models suffer from poor loss landscape geometry compared to unconstrained models, even when there is an equivariant optimal solution \cite{xie_tale_2025, nordenfors_optimization_2024}.
These findings suggest that strict constraints throughout training can obstruct convergence and limit final performance.

To mitigate these issues, many studies have focused on relaxed or approximate equivariance.
For example, \citet{wang_approximately_2022} break the weight sharing scheme in group convolutions.
Other methods use non-equivariant residuals \cite{finzi_residual_2021}, learned partial equivariances \cite{romero_learning_2023}, non-stationary filters \cite{ouderaa_relaxing_2022}, projection regularization \cite{berndt_approximate_2026}, probabilistic equivariance \cite{veefkind_probabilistic_2025}, or approximate-equivariance objectives \cite{elhag_relaxed_2026}.
More related to our work is symmetry breaking through auxiliary geometric reference information \cite{vadgama_probing_2025}.
\blue{
However, such persistent symmetry breaking changes the deployed model, which is no longer equivariant at inference.
It can therefore exploit dataset-specific biases and increase vulnerability to distribution changes, whereas our method breaks the symmetry only during training and deploys the original strictly equivariant model.}
Equivariance can also be imposed on non-equivariant architectures by group averaging \cite{puny_frame_2022} or canonicalization \cite{kaba_equivariance_2023}, often approximately, for example by averaging over a subgroup for efficiency. 

More recently, studies have focused on relaxation specifically to aid optimization. 
Pertigkiozoglou et al.~\citep{pertigkiozoglou_improving_2025, pertigkiozoglou_recurrent_2026} and \citet{manolache_learning_2025} have introduced methods that add non-equivariant dense layers during training, which are annealed to zero to recover the strictly equivariant architecture at inference.
These methods show that traversing a non-equivariant optimization path can improve the convergence of equivariant models.
However, as we show here (e.g., in \Cref{tab:pointcloud-results}), such methods significantly increase parameter count and training time,
making them computationally prohibitive at scale.

\textbf{Positional Encodings in Equivariant Models.}
Positional encodings are widely used to inject structural information into architectures that are otherwise symmetry-preserving, most notably in Transformers \cite{vaswani_attention_2017, dosovitskiy_image_2020}. 
In geometric deep learning, positional encodings have been employed to distinguish indistinguishable nodes in GNNs \cite{wang_equivariant_2022, dwivedi_graph_2022}, and are known to strengthen expressivity and restore universality in both Transformers and message passing networks \cite{yun_are_2020, abboud_surprising_2021, sato_random_2021}. \\
Recent studies have begun to leverage auxiliary symmetry-breaking inputs or objects specifically as a tool for controlling symmetry properties. 
\citet{goel_any-subgroup_2025} utilize positional information to implement generic equivariant models that can relax constraints to specific subgroups, \citet{lawrence_improving_2025} employ randomized positional encodings to implement probabilistic symmetry breaking, and \citet{ashman_approximately_2024} construct approximately equivariant models from existing equivariant architectures through an auxiliary-input mechanism. 
Our approach builds upon this direction but focuses on the specific application of annealing these positional encodings to facilitate the training of strictly equivariant models, serving as a parameter-efficient alternative to dense-layer relaxation.

\vskip -0.05 in
\section{Preliminaries} \label{sec:preliminaries}
\vskip -0.05 in

We consider learning tasks where a symmetry group naturally characterizes the data. 
Let $G$ be a compact group acting on an input space $\mathcal{X}$ and an output space $\mathcal{Y}$ via representations $\rho_{\mathrm{in}}: G \rightarrow \mathrm{GL}\left(\mathcal{X}\right)$ and $\rho_{\mathrm{out}}: G \rightarrow \mathrm{GL}\left(\mathcal{Y}\right)$. 
A function $f: \mathcal{X} \rightarrow \mathcal{Y}$ is said to be $G$-\emph{equivariant} if it commutes with the group action. 
That is, for all $g  \in G$ and $x \in \mathcal{X}$:
\inlineeq{
    f \left(\rho_{\mathrm{in}} \left( g \right)  x \right) = \rho_{\mathrm{out}} \left( g \right) f \left( x \right) \,.
    } \\
When the output space is invariant under the group action (e.g., in classification tasks where the label is unchanged by rotation), this condition simplifies to $f \left( \rho_{\mathrm{in}} \left( g \right) x \right) = f \left( x \right)$, and the function is termed \emph{invariant}.
To simplify notation, we use the shorthand $g\cdot(\,\cdot\,)$ to denote the appropriate group action (e.g., $g \cdot x = \rho_{\mathcal{X}} \left( g \right) x $) whenever it is clear from context.

\textbf{Equivariant Neural Networks.}
Equivariant neural networks (ENNs) typically enforce this property by restricting the hypothesis class to compositions of standard equivariant (often linear) maps and pointwise nonlinearities \cite{zaheer_deep_2017,maron_invariant_2018}. 
Concretely, 
\inlineeq{
f_\theta^\mathrm{eq} = f_{\theta}^{\mathrm{eq},L} \circ \sigma \circ \dots \circ \sigma \circ f_{\theta}^{\mathrm{eq},1}\,,
} \\
where an equivariant layer $f_{\theta}^{\mathrm{eq},i}$ satisfies
\inlineeq{
    f_{\theta}^{\mathrm{eq},i} \left( \rho_{\mathrm{in}}\left( g \right)  x \right)    = \rho_{\mathrm{out}}  \left( g \right) f_{\theta}^{\mathrm{eq},i} \left( x \right) \,, \forall g \in G \,.
    }

\textbf{Symmetry breaking via relaxation.}
Strict equivariance provides a strong inductive bias but may be overly restrictive, for example, when symmetries are only approximate \cite{wang_approximately_2022} or when the task requires breaking self-symmetries \cite{lawrence_improving_2025}.
\citet{finzi_residual_2021} have proposed to soften equivariance by augmenting each equivariant layer with a non-equivariant residual pathway.
More recent works repurpose this idea for train-time constraint relaxation \cite{pertigkiozoglou_improving_2025,manolache_learning_2025,pertigkiozoglou_recurrent_2026}, using
$f_{\mathrm{relaxed}}(x) = f_{\theta}^{\mathrm{eq},i}(x) + \lambda f_{\phi}^{\mathrm{neq},i}(x)$,
where $f_{\phi}^{\mathrm{neq},i}$ is an unconstrained map, such as a dense linear layer \cite{pertigkiozoglou_improving_2025} or a multi-layer perceptron (MLP) \cite{manolache_learning_2025,pertigkiozoglou_recurrent_2026}.
The coefficient $\lambda$ controls the relaxation strength and is driven to zero during training, recovering strict equivariance at inference.
However, the auxiliary non-equivariant branches increase parameter count and training cost, which can become prohibitive when the equivariant layer is parameter-efficient.
For example, a dense map over flattened high-resolution feature maps scales with the global input size, making it far more expensive than a local equivariant convolution.

\textbf{Positional Encodings.}
Positional encodings (PE) inject order or geometry into otherwise permutation-invariant architectures, e.g., sinusoidal embeddings that break the permutation invariance of attention \citep{vaswani_attention_2017}.
\blue{
Throughout the analysis, we take $\mathcal{P}$ to be a compact $G$-invariant subset of an ambient normed representation space with a linear $G$-action.
Thus, for $\lambda p\in\mathcal{P}$, we have $g\cdot(\lambda p)=\lambda(g\cdot p)$ and, for $\lambda>0$, $\mathrm{Stab}(\lambda p)=\mathrm{Stab}(p)$.
}
We distinguish between two PE integration strategies $\phi: \mathcal{X} \times \mathcal{P} \to \mathcal{Z}$: \\
\emph{Concat-based PE.} The input $x$ is augmented with positional features $p$ along new channel dimensions, effectively lifting the input to a higher-dimensional joint space: 
$     \phi_{\mathrm{cat}} \left(x, p \right) = \left[ x \parallel p \right] \,.$
In this case, we view $(x,p)$ as an element of $X \times P$ endowed with product norm $\|(x,p)\|_{\mathcal{X} \times \mathcal{P}} = \|x\|_\mathcal{X} + \|p\|_\mathcal{P}$, and the group acts jointly as $g \cdot (x,p) := (g \cdot x, g \cdot p)$.
This preserves the distinction between features and positional encodings, as in GNN methods such as PEG \cite{wang_equivariant_2022} and LSPE \cite{dwivedi_graph_2022} that distinguish otherwise indistinguishable nodes. \\
\emph{Additive PE.} The positional features are added directly to the input signal: $\phi_{\mathrm{add}} \left( x, p \right) = x + p \,.$
Here we assume that $x$ and $p$ lie in a common ambient representation space, or are first embedded into one, so that addition and the group action are defined on both terms.
This is a common approach in Transformers \cite{vaswani_attention_2017, dosovitskiy_image_2020}, where sinusoidal or learnable position vectors are added to token embeddings.

\vskip -0.05 in
\section{Constraint relaxation via Positional Encoding}
\label{sec:method_poe}
\vskip -0.05 in

In this work, we propose a parameter-efficient alternative to constraint relaxation with dense layers: symmetry breaking via Positional Encoding.
Instead of augmenting the equivariant architecture with auxiliary dense layers, we modify the input representation itself. \\
Let $f_{\theta} : \mathcal{X} \times \mathcal{P} \to \mathcal{Y}$
be a neural network that is strictly $G$-equivariant with respect to the joint action $g \cdot \left( x,p \right) := \left( g \cdot x,\  g \cdot p \right)\,, g \in G $.
Let $p_0 \in \mathcal{P}$ be an anchor positional encoding and $\lambda \in [0,1]$ a scalar relaxation coefficient.
We define the relaxed model as
$
h_{\lambda} \left(x \right) := f_{\theta} \left( x,\lambda p_0 \right) \,.
$ \\
During training we have $\lambda > 0$, and the anchor $p_{0}$ breaks global $G$-equivariance, whereas in the limit $\lambda \to 0$ the input becomes $(x,0)$, restoring the equivariance of the original architecture.
Thus, $\lambda$ induces a transition from a rich non-equivariant hypothesis space to the strictly equivariant subspace, similar to \citet{pertigkiozoglou_improving_2025} and \citet{manolache_learning_2025}.
Our analysis below focuses on concat-based PE: we link it to the symmetry-breaking expressivity principle of \citet{goel_any-subgroup_2025} and derive a vanishing equivariance-error bound analogous to the dense-layer bounds of \citet{manolache_learning_2025}.
We then discuss how the same principle motivates additive PE.

\textbf{Universality of the relaxed model.}
Although the mechanism behind the benefits of constraint relaxation remains unclear, the approach is often motivated as temporarily enlarging the hypothesis class, so that optimization can pass through less constrained, approximately equivariant models before returning to the equivariant subspace \cite{pertigkiozoglou_improving_2025, manolache_learning_2025}. 
This suggests a basic desideratum: the relaxed model should be strictly more expressive than the constrained one. 
For dense-layer relaxation, this is immediate, since the added unconstrained layers yield an MLP-like universal approximator \cite{hornik_multilayer_1989}. 
Below, we show that our PE-based relaxation provides an analogous universality result while adding only a few parameters.

\begin{proposition}
\label{prop:universality-via-pe}
\blue{Let $G$ be a compact group acting on compact spaces $\mathcal{X}$ and $\mathcal{P}$, and let $\mathcal{Y}$ be a normed vector space.}
Let \blue{a family of networks} $f_{\theta}:\mathcal{X} \times \mathcal{P} \rightarrow \mathcal{Y}$ be a universal approximator over the continuous G-equivariant functions on the product space $\mathcal{X} \times \mathcal{P}$.
Let $p_0 \in \mathcal{P}$ be a fixed element with no self-symmetries in $G$, i.e. $g\cdot p_0 = p_0 \Rightarrow g=e$, and let $\lambda > 0$.
Then, the family of functions $h(\cdot) = f_\theta(\cdot, \lambda p_0): \mathcal{X} \rightarrow \mathcal{Y}$ obtained by fixing the second argument of a universal $G$-equivariant network is a universal approximator for the space of continuous functions $C(\mathcal{X}, \mathcal{Y})$.
\end{proposition}
\vskip -0.05 in
\begin{proof}
\blue{
The result follows from Theorem 2 of \citet{goel_any-subgroup_2025}, which characterizes the expressivity of networks conditioned on symmetry-breaking features. The Theorem states that for a $G$-equivariant network $f(x,p)$, fixing the auxiliary input $p$ to a value $p_0$ restricts the network's equivariance to the subgroup $\mathrm{Stab}(p_0) \leq G$, while remaining a global approximator under this weaker constraint. 
Since $p_0$ has no self-symmetries, i.e., $\mathrm{Stab}(p_0)=\{e\}$, and since $\lambda>0$ implies $\mathrm{Stab}(\lambda p_0)=\mathrm{Stab}(p_0)$, $h$ becomes equivariant only to the identity element, which imposes no constraints. Therefore, the hypothesis space of $h$ expands to include all continuous functions, following the universal approximation theorem~\citep{hornik_multilayer_1989}.
}
\end{proof}
This proposition motivates our method: for $\lambda>0$, optimization can traverse a universal, generally non-equivariant hypothesis class before annealing back to the equivariant class at $\lambda=0$.

\textbf{Equivariance error bound.}
Although our relaxed model is non-equivariant during training, we can bound its deviation from strict equivariance, as in the dense-layer case of \citet{manolache_learning_2025}.
Such a limited deviation may be beneficial, as has been shown by \citet{pertigkiozoglou_improving_2025}, where explicitly regularizing the equivariance error of the auxiliary dense layers is crucial.

\begin{proposition} \label{prop:eq-error-bound}
    \blue{Let \(f_\theta=f^L\circ\cdots\circ f^1:\mathcal X\times\mathcal P\to\mathcal Y\)
    be strictly \(G\)-equivariant with respect to the joint action
    \(g\cdot(x,p)=(g\cdot x,g\cdot p)\), and suppose that each layer \(f^\ell\) is
    \(M\)-Lipschitz continuous as a map between its corresponding input and output
    representation spaces.}
    Assume that the group action is isometric.
    The equivariance error of the relaxed model $h_{\lambda}(x) = f_\theta \left( x, \lambda p_0 \right)$ is bounded by:
    \begin{equation*}
    \lVert h_{\lambda} \left( g \cdot x \right) - g \cdot h_{\lambda}\left( x \right) \rVert \leq  2 \lambda M^L \lVert p_0 \rVert
    \end{equation*}
\end{proposition}

The proof is provided in \Cref{sec:equivariance-error-bound-proof}.
The bound implies that annealing $\lambda \to 0$ drives the equivariance violation to zero at least linearly in $\lambda$, with the remaining constants determined by the anchor magnitude $\|p_0\|$ and the network's Lipschitz growth.
In contrast, the bound of \citet{manolache_learning_2025} scales with the relaxation coefficients and typically depends on $\|x\|$, whereas here the deviation is controlled directly by $\lambda\|p_0\|$ (up to amplification by the Lipschitz constant).

\textbf{Additive Positional Encoding.}
In many modern architectures, such as Transformers, it is standard practice to use an additive PE. 
In our relaxed-model notation, this takes the form $h_{\lambda}(x) = f_{\theta} \left( x + \lambda p_0 \right)$. 
In this case, the equivariance error bound from \Cref{prop:eq-error-bound} applies directly.
However, proving universality is more challenging here than in the concatenative case: the general guarantee from \Cref{prop:universality-via-pe} does not automatically hold and requires specific structural assumptions.
Nevertheless, theoretical precedents confirm that additive encodings can indeed restore universality to equivariant models.
For example, \citet{yun_are_2020} prove that permutation equivariant Transformers with distinct positional encodings become universal approximators for general sequence-to-sequence functions.
Similarly, \citet{abboud_surprising_2021} show that augmenting equivariant Message Passing Networks with unique node identifiers is sufficient to restore universality.

\vskip -0.05 in
\section{Experiments} \label{sec:experiments}
\vskip -0.05 in
We implement our proposed framework across five distinct architectures.
As in previous train-time symmetry-breaking studies \citep{pertigkiozoglou_improving_2025, manolache_learning_2025}, we evaluate our method on the Vector Neurons framework \cite{deng_vector_2021} and SEGNN \cite{brandstetter_geometric_2022}.
We also evaluate our method on the more recent Alias-Free XCiT \cite{michaeli_alias-free_2025} and Rapidash \cite{vadgama_probing_2025}, and on E2CNN \cite{weiler_general_2019} for the smoke plume dataset \citep{wang_approximately_2022}, where we additionally compare against 
approaches that keep the symmetry breaking at inference.
In all cases, we introduce a positional encoding weighted by a learnable scale $\gamma$ initialized to $\gamma=1$, and a scheduling coefficient $\lambda$ that decays to 0, recovering strict equivariance at inference.
\blue{
Here the bound of \Cref{prop:eq-error-bound} applies with $p_0=\gamma p$, whose magnitude is controlled by the regularizer on $\|\lambda(t)\gamma p\|_2$ introduced below.
}
\vskip -0.05 in
\subsection{Rotation-Equivariant Point Cloud Networks}
\vskip -0.05 in
\textbf{Vector Neurons.}
We apply our method to \textit{VN-PointNet} and \textit{VN-DGCNN}, built upon the Vector Neurons framework \cite{deng_vector_2021} for point cloud classification. 
Based on the original $S_n$ equivariant \textit{PointNet} \cite{Qi2016} and \textit{DGCNN} \cite{wang_dynamic_2019}, 
these models replace scalar neurons with 3D vector neurons that rotate together with the input, yielding $S_n \times SO(3)$ equivariance. \\
We break this rotation symmetry by introducing a global positional encoding $p \in \mathbb{R}^{N \times 3}$ that remains fixed in the canonical coordinate frame regardless of the input rotation. 
We examine two variants: \emph{Fixed-PE}, consisting of a single learnable $\mathbb{R}^{3}$ vector broadcast across all $N$ points, 
and a \emph{Sinusoidal-PE}, a deterministic per-index encoding following \citet{vaswani_attention_2017},
concretely, 
\inlineeq{p_{k, 2i} = \sin\left(\frac{k}{10000^{2i/3}}\right),\, p_{k, 2i+1} = \cos\left(\frac{k}{10000^{2i/3}}\right) \label{eq:sin-pe} \,,} \\
where $k \in \{1, \dots, N\}$ is the point index and $2i, 2i+1 \in \{0, 1, 2\}$ index the coordinates of $p_k \in \mathbb{R}^3$. \\
Fixed-PE breaks $SO(3)$ through a broadcast reference vector,\footnote{\blue{
Note Fixed-PE has a non-trivial \(SO(3)\) stabilizer: rotations around the broadcast vector.
We discuss this caveat and a possible solution in \Cref{app:fixed-pe-stabilizer}.
}}
whereas Sinusoidal-PE also relaxes the $S_n$ permutation symmetry through unique per-index encodings.
This differs from the dense-layer relaxation \citep{pertigkiozoglou_improving_2025}, which breaks $SO(3)$ while preserving permutation symmetry.
We inject PE either additively, $x' = x + \lambda \gamma p \in \mathbb{R}^{N \times 3}$, or by concatenation, $x' = [x, \lambda \gamma p] \in \mathbb{R}^{N \times 6}$ for $x \in \mathbb{R}^{N \times 3}$.

\textbf{Rapidash.}
\blue{
We further evaluate our method on \emph{Rapidash} \citep{vadgama_probing_2025}, a regular group-convolution architecture for point clouds defined over the position-orientation space \(\mathbb{R}^3 \times S^2\).
Following \citet{vadgama_probing_2025}, we consider the strongest ModelNet40 variants: (i) \emph{Rapidash-Eq}, which uses point-cloud normals as vector features and remains strictly equivariant at test time, and (ii) \emph{Rapidash-Symm.~Break}, which additionally uses global-frame features and therefore breaks the symmetry also at inference.
On top of the strictly equivariant baseline, we evaluate \emph{Rapidash-Eq+PE}, obtained by injecting the PE on the input similarly to the Vector Neurons.
This positional signal breaks rotational symmetry during training, and annealing it to zero recovers the original strictly equivariant Rapidash model at inference.
}
\vskip -0.05 in
\subsection{Equivariant Models on 2D Grids}
\vskip -0.05 in
We consider two symmetry groups acting on data defined over 2D grids: fractional cyclic translations of natural images, and planar rotations of fluid velocity fields.
In both cases, we break the symmetry using a 2D Fourier positional encoding of the grid coordinates.

\textbf{Fractional shift-invariant vision transformers.}
We evaluate our method on the \textit{Alias-Free Cross-Covariance Image Transformer} (AF-XCiT; \citet{michaeli_alias-free_2025}).
We consider the symmetry group of fractional cyclic translations, formally the torus group $G \cong \mathbb{R}^2 / \mathbb{Z}^2$, acting on the periodic continuous image domain as defined by \citet{michaeli_alias-free_2023}.
AF-XCiT achieves translation invariance\footnote{
In practice, AF-XCiT is only approximately equivariant due to imperfect alias-free activation functions, compared to the complete alias-free case in \citet{michaeli_alias-free_2023} where equivariance is guaranteed.}
by modifying the XCiT architecture \cite{el-nouby_xcit_2021} using alias-free downsampling and activations, translation-equivariant cross-covariance attention layers, and no positional encodings. \\
To implement constraint relaxation, we reintroduce the 2D Fourier Positional Encoding used in the original XCiT \cite{el-nouby_xcit_2021} after the patch embedding and prior to the first transformer block.
This module encodes the patch coordinates with sinusoidal features and linearly projects them to the model feature dimension \(d\).
This encoding injects absolute position into the image features, formally:
\inlineeq{
    \mathrm{PatchEmbed}: \mathbb{R}^{H\times W \times 3} \rightarrow \mathbb{R}^{N \times d}\,; \quad
    z' = \mathrm{PatchEmbed} \left( x \right) + \lambda \gamma P_{\text{Fourier}} \,,
} \\
where $x$ is the input image with dimensions $H,W$ and $P_{\text{Fourier}} \in \mathbb{R}^{N \times d}$ is the positional signal.

\textbf{Rotation-equivariant steerable CNNs.}
We evaluate our method on the synthetic smoke plume dataset with approximate rotation symmetry of \citet{wang_approximately_2022}, where the task is to forecast 2D smoke velocity fields simulated with inflow locations and buoyant force directions that are $C_4$-symmetric, while the buoyancy factor varies with the inflow position and breaks this symmetry.
Our strict baseline is the $C_4$-equivariant steerable CNN E2CNN \citep{weiler_general_2019}, which we relax during training using the 2D Fourier positional encoding described above, scaled by $\lambda \gamma$.
We compare against RSteer \citep{wang_approximately_2022}, which relaxes the steerable-kernel constraint and deploys the resulting approximately equivariant model at inference.

\vskip -0.05 in
\subsection{Steerable E(3)-Equivariant GNNs}
\vskip -0.05 in
Finally, we evaluate our approach on \emph{SEGNN} \cite{brandstetter_geometric_2022} on the charged N-Body task \cite{kipf_neural_2018}. 
While SEGNN is often described as \(E(3)\)-equivariant, in our setting the relevant symmetry is global rotations (\(SO(3)\)), since global translations are handled by the standard use of centered coordinates.
SEGNN preserves the symmetry of input features represented by a direct sum of irreducible representations (irreps) of the orthogonal group \(O(3)\).
This is suitable for physical simulations where the inputs consist of multiple steerable vector features, such as relative position and velocity, and invariant scalar features, such as energy. \\
\blue{
To implement symmetry breaking, we apply additive PE to the full SEGNN input feature vector:
$x' = x + \lambda \gamma p \,, x,p \in \mathbb{R}^{N \times D}\,,$
where \(D=\sum_{\ell \in \mathcal{L}_{\mathrm{in}}} d_\ell(2\ell+1)\) is the total flattened feature dimension induced by the input irreps. 
Thus, \(p\) has the same irrep-flattened layout as the SEGNN input features.
We instantiate \(p\) either by broadcasting a learnable vector in \(\mathbb{R}^{D}\) across nodes (Fixed-PE), or by applying the sinusoidal encoding (\Cref{eq:sin-pe}) to the node index with feature dimension \(D\) (Sinusoidal-PE).
This positional signal is fixed in the canonical feature basis and is not transformed with the input, thereby breaking rotational symmetry during training; annealing \(\lambda\) to zero recovers the original SEGNN representation at inference.
}

\textbf{Annealing schedules.}
As mentioned in \Cref{sec:method_poe}, our PE constraint relaxation during training is controlled by $\lambda(t)$, where $t$ represents the current training step out of $T$ total steps.
We examine two scheduling strategies. \\
\emph{Linear Decay.} A monotonic schedule that initializes the model with maximal relaxation ($\lambda=1$), linearly annealing to strict equivariance as training concludes, formally  \inlineeq{\lambda_{\text{Linear}}(t) = 1 - \frac{t}{T}\,.} \\
\emph{Triangular Schedule.} A cyclic schedule that starts and ends with strict equivariance ($\lambda=0$), with maximal relaxation midway through training, as first proposed by \citet{pertigkiozoglou_improving_2025}. 
This schedule allows the model to temporarily explore non-equivariant solutions before converging back to the equivariant subspace, formally \inlineeq{ \lambda_{\text{Triangular}}(t) = 
    \begin{cases} 
        \frac{2t}{T} & \text{if } t < \frac{T}{2} \\
        2 - \frac{2t}{T} & \text{if } t \geq \frac{T}{2}
    \end{cases} \,.} \\

\textbf{Training objective.}
We train all models with their original task loss $\mathcal{L}_{\text{task}}$  combined with a regularization term on the magnitude of the injected PE:
\inlineeq{\mathcal{L} = \mathcal{L}_{\text{task}} + \beta \cdot \|\lambda(t) \gamma p\|_2 \,,} \\
where $\beta$ is a regularization hyperparameter. 
This penalizes the norm of the symmetry-breaking signal, encouraging the model to rely less on the positional cues as training progresses.
This is simpler than \citet{pertigkiozoglou_improving_2025}, which regularizes both the magnitude and equivariance error of dense relaxation layers.

\vskip -0.05 in
\section{Results}\label{sec:results}
\vskip -0.05 in
We run all experiments using PyTorch \citep{paszke_automatic_2017} on a single machine with 8 \(\times\) NVIDIA RTX A6000.
We build on the official codebases of previous studies for all experiments \citep{michaeli_alias-free_2025, pertigkiozoglou_improving_2025, vadgama_probing_2025, wang_approximately_2022}.
Our implementation is publicly available,\footnote{\url{https://github.com/hmichaeli/pe_constraint_relaxation}} and further experimental details are provided in \Cref{sec:app-experimental-details}.
\vskip -0.05 in
\subsection{Rotation-Equivariant Point Cloud Networks} \label{sec:results-point-clouds}
\vskip -0.05 in
\textbf{Vector Neurons.}
We evaluate our PE relaxation on the ModelNet40 point-cloud classification task \cite{wu_3d_2015}.
We compare against two baselines: strictly equivariant Vector Neurons models (VN-PointNet and VN-DGCNN) \cite{deng_vector_2021} and the dense-layer Constraint Relaxation (CR) method \citep{pertigkiozoglou_improving_2025}.
Following \citet{pertigkiozoglou_improving_2025}, we sample 300 points per point-cloud and keep the same training hyperparameters, tuning only our framework-specific choices (regularization scale, mix coefficient scheduler, and positional encoding type), which we tune as reported in \Cref{sec:app-point-cloud}. 
For fairness, we run the same hyperparameter search for dense-layer CR and report the best configuration averaged over 5 seeds. 
All VN experiments use the $SO(3)/SO(3)$ protocol where point clouds are randomly rotated during both training and evaluation \citep{deng_vector_2021}. 

The results in \Cref{tab:pointcloud-results} show that our PE method outperforms the equivariant baseline and the CR method on the VN-PointNet architecture, with relative improvements of 18\% and 7.6\%, respectively.
On VN-DGCNN, our method matches CR accuracy without the high cost of dense relaxation.
Beyond accuracy, \Cref{tab:pointcloud-results} shows the computational advantages of our method. 
Across both VN architectures, PE preserves almost the same parameter count and training time as the equivariant baseline, whereas dense-layer CR incurs substantial overhead. 
For example, on VN-PointNet, additive PE improves accuracy from \(67.5\) to \(79.6\) with nearly unchanged training cost (\(2\)M parameters and \(27.9\)s per epoch, versus \(2\)M and \(27.6\)s for the baseline), whereas CR increases the parameter count to \(9\)M  and training time to \(97.6\)s per epoch. 
A similar trend appears for VN-DGCNN.

We report additional scalability results in \cref{app:vn_scalability}, which confirm that PE continues to track the baseline as the model scales, while CR exhibits rapidly growing overhead. 
Additional ablations of the PE design choices in \Cref{sec:app-ablation-vn} show that the learnable scale $\gamma$ accounts for most of the gain of our method: learning $\gamma$ without decay or regularization already yields comparable accuracy after PE removal, whereas with a fixed $\gamma$ removing PE collapses accuracy. 
The schedule $\lambda(t)$ mainly guarantees that the PE contribution is exactly zero at the end of training.

\begin{table*}[t]
  \caption{\textbf{ModelNet40 classification accuracy and training cost.} Results for VN-PointNet and VN-DGCNN under strict equivariance, dense-layer constraint relaxation (CR), and our positional-encoding relaxation (PE; additive and concatenative variants). Accuracy is the average best validation accuracy over training (mean $\pm$ std over 5 random seeds). 
  All models are trained and evaluated on randomly rotated point clouds ($SO(3)/SO(3)$).
  ``Parameters'' counts trainable parameters in the trained architecture, and ``Time per epoch'' is computed by dividing total wall-clock training time on a single NVIDIA RTX A6000 with batch-size 32 by the number of epochs. }
  \label{tab:pointcloud-results}

  \begin{center}
    \begin{small}
      \begin{sc}
        \begin{tabular}{lcccr}
          \toprule
    Model   & \doublelinecell[]{Instance \\ Accuracy}  & Parameters & \doublelinecell[]{Time per\\Epoch [sec]}\\
    \midrule
    VN-PointNet \cite{deng_vector_2021}                     & 67.5 $\pm$ 0.5           & 1,972,745  & 27.6  \\
    VN-PointNet-CR \cite{pertigkiozoglou_improving_2025}    & 74.0 $\pm$ 1.9           & 8,978,126  &  97.6 \\
    VN-PointNet-PE add. (ours)                             & \textbf{79.6} $\pm$ 0.4  & 1,972,746  &  27.9
  \\
    VN-PointNet-PE concat. (ours)                          & 79.3 $\pm$ 0.9           & 1,972,872  &  27.7  \\
    \midrule
    VN-DGCNN \cite{deng_vector_2021}                        & 88.3 $\pm$ 0.2           & 2,899,830  &  61.2  \\
    VN-DGCNN-CR \cite{pertigkiozoglou_improving_2025}       & 88.6 $\pm$ 0.5           & 8,829,555  &  181.7   \\
    VN-DGCNN-PE add. (ours)                                & 88.2 $\pm$ 0.2           & 2,899,834  &  61.3  \\
    VN-DGCNN-PE concat. (ours)                             & 88.6 $\pm$ 0.4           & 2,899,915  &  61.7  \\
    \bottomrule
        \end{tabular}
      \end{sc}
    \end{small}
  \end{center}
  \vskip -0.1in
\end{table*}

\textbf{Rapidash.}
We next evaluate our method on Rapidash on ModelNet40.
Similar to Vector Neurons, we use the original model hyperparameters \cite{vadgama_probing_2025} and tune those of our framework as described in \Cref{sec:app-rapidash}.
We report in \Cref{tab:rapidash-results} the accuracy on the original validation set, and validation accuracy under random rotations.
For a controlled comparison on the same backbone, we also implement the constraint relaxation of \citet{pertigkiozoglou_improving_2025} on Rapidash-Eq (\emph{Rapidash-CR}; details in \Cref{sec:app-rapidash}).
Here we follow the protocol of~\cite{vadgama_probing_2025}: training uses the original, $z$-aligned ModelNet40 with 1024 points, and each training sample is randomly rotated inside the model pipeline. 
\emph{Rapidash-Symm.~Break} receives this applied rotation as a global-frame pose feature, while \emph{Rapidash-Eq}, \emph{Rapidash-CR and Rapidash-Eq+PE} receive no pose information at inference. 
``Accuracy'' uses the aligned validation set with the identity pose; ``Random-Rotation Accuracy'' applies an additional rotation that is not encoded in the pose feature, which is why Rapidash-Symm.~Break collapses.\\
The results reveal a clear dependence of the global-frame variant on the orientation bias in the dataset.
While \emph{Rapidash-Symm.~Break} achieves the highest standard validation accuracy (88.0), its performance drops sharply under random rotations to 14.8.
In contrast, the strictly equivariant \emph{Rapidash-Eq} baseline remains much more stable (83.7 to 83.6), while our \emph{Rapidash-EQ+PE} improves over this baseline on both metrics, reaching 84.6 accuracy on both the standard and the randomly rotated validation sets.
Rapidash-CR likewise improves over the baseline and remains stable under rotations (84.3 and 84.2), while PE performs slightly better.
These results suggest that PE provides the optimization benefit of symmetry breaking during training without causing the model to rely on dataset-specific orientation bias at test time.
By annealing the positional signal to zero, \emph{Rapidash-Eq+PE} recovers the original equivariant architecture while retaining improved performance.

\begin{table}[t]
  \caption{
  \textbf{ModelNet40 classification with Rapidash.}
  We report validation accuracy on the original validation set (``Accuracy'') and on the validation set under random rotations (``Random-Rotation Accuracy'').
  PE improves both standard accuracy and rotated accuracy over the strictly equivariant Rapidash baseline, while preserving the original equivariant test-time architecture.
  In contrast, Rapidash-Symm. Break achieves the highest standard accuracy but collapses under random rotations, indicating strong reliance on the orientation bias present in the dataset.
  }
  \label{tab:rapidash-results}
  \begin{center}
    \begin{small}
      \begin{sc}
      \begin{tabular}{lcc}
        \toprule
        Model & Accuracy & \doublelinecell[]{Random-Rotation\\Accuracy} \\
        \midrule
        Rapidash-Eq \cite{vadgama_probing_2025} & 83.7 $\pm$ 1.4 & 83.6 $\pm$ 1.0  \\
        Rapidash-Symm. Break \cite{vadgama_probing_2025} & \textbf{88.0} $\pm$ 0.1 & 14.8 $\pm$ 1.1 \\
        Rapidash-CR \cite{pertigkiozoglou_improving_2025} & 84.3 $\pm$ 0.8 & 84.2 $\pm$ 0.5 \\
        Rapidash-Eq+PE (ours) & 84.6 $\pm$ 0.6  & \textbf{84.6} $\pm$ 0.5 \\
        \bottomrule
      \end{tabular}
      \end{sc}
    \end{small}
  \end{center}
  \vskip -0.1in
\end{table}

\vskip -0.05 in
\subsection{Equivariant Models on 2D Grids}
\vskip -0.05 in
\textbf{Fractional shift-invariant vision transformers.}
We evaluate our method on AF-XCiT-Nano \cite{michaeli_alias-free_2025}, trained on ImageNet \cite{deng_imagenet_2009}. 
We use the original model hyperparameters and set the regularization term to $\beta = 10^{-5}$.
We use linear relaxation decay for 320 epochs, and train the remaining 80 epochs with $\lambda = 0$ (see additional details in \Cref{sec:app-xcit}).
AF-XCiT may be an instance where the model symmetry is too restrictive, as natural images are not cyclically shift-symmetric (e.g., due to boundary effects).
To examine this hypothesis, we train an additional AF-XCiT variant with a non-decaying positional encoding, i.e., where the PE breaks equivariance at test time.
The results in \Cref{tab:af-xcit} suggest that our proposed approach improves the accuracy of the AF model (71.1\% vs 70.6\%).
The model with non-decaying PE performs slightly better (71.4\%), supporting our hypothesis regarding the symmetry misalignment.
Crucially, however, our proposed approach achieves this gain without compromising the translation consistency, which is beneficial, e.g., for adversarial robustness---the original motivation for AF-XCiT \cite{michaeli_alias-free_2025}. We discuss this further in \Cref{sec:app-adversarial-robustness}.

\begin{table*}
\caption{
  \textbf{ImageNet results for AF-XCiT-Nano.} 
  We report ImageNet validation top-1 accuracy (``Accuracy''), accuracy under random cyclic translations (``Shift Accuracy''), and the relative change between the two. 
  We report mean $\pm$ std over 2 random seeds.
  AF-XCiT + PE (train and test) keeps positional encoding at inference and therefore corresponds to a relaxed, only approximately invariant setting, whereas AF-XCiT + PE (train only; ours) anneals the PE coefficient to zero during training and removes PE at test time, recovering the original alias-free invariant architecture. 
  The results show that the annealed variant preserves substantially better accuracy under input translations than keeping PE at test time, while still improving accuracy over the strict AF-XCiT baseline.
}

  \label{tab:af-xcit}
  \begin{center}
    \begin{small}
      \begin{sc}
      \begin{tabular}{lccc}
        \toprule
    Model & Accuracy     &  Shift accuracy & Change (\%) \\
    \midrule
    XCiT-Nano \cite{el-nouby_xcit_2021}                & 70.5 $\pm$ 0.2 & 68.8 $\pm$ 0.0 & -2.4 \\
    AF-XCiT-Nano \cite{michaeli_alias-free_2025}       & 70.6 $\pm$ 0.1 & 70.6 $\pm$ 0.1 & -0.0 \\
    AF-XCiT-Nano + PE (train and test)                 & \textbf{71.4} $\pm$ 0.1 & 70.3 $\pm$ 0.1 & -1.5 \\
    AF-XCiT-Nano + PE (train only; ours)               & 71.1 $\pm$ 0.1 & \textbf{71.1} $\pm$ 0.1 & -0.0 \\
    \bottomrule

  \end{tabular}
      \end{sc}
    \end{small}
  \end{center}
  \vskip -0.1in
\end{table*}

\textbf{Rotation-equivariant steerable CNNs.}
\Cref{tab:smoke} reports the 20-step prediction RMSE on the smoke plume dataset with approximate rotation symmetry \citep{wang_approximately_2022}, for the Future (test data lies in the future time of the training data) and Domain (training and test data are from different simulations) test sets.
Both PE variants reduce the RMSE of the strict E2CNN baseline on both test sets.
RSteer obtains the lowest RMSE, while PE (train only) is within $0.4$ of RSteer on both test sets. 
Note that the methods target different deployment goals: RSteer and PE (train and test) deploy an approximately equivariant model, whereas PE (train only) recovers the strictly $C_4$-equivariant E2CNN at inference.
Decayed PE thus comes close to RSteer while remaining strictly equivariant.

\begin{table}[t]
  \caption{
  \textbf{Smoke plume forecasting with E2CNN.}
  We report 20-step prediction RMSE ($\times 10^{-2}$, lower is better) on the Future and Domain test sets of \citet{wang_approximately_2022} (mean $\pm$ std over 3 seeds).
  E2CNN + PE (train and test) keeps the PE at inference and is therefore only approximately equivariant, similar to RSteer, whereas E2CNN + PE (train only; ours) anneals the PE to zero during training and recovers the strictly equivariant E2CNN at inference.
  }
  \label{tab:smoke}
  \begin{center}
    \begin{small}
      \begin{sc}
      \begin{tabular}{lcc}
        \toprule
        Model & Future $\downarrow$ & Domain $\downarrow$ \\
        \midrule
        E2CNN \cite{weiler_general_2019} & 81.2 $\pm$ 1.1 & 59.8 $\pm$ 2.0 \\
        RSteer \cite{wang_approximately_2022} & \textbf{79.4} $\pm$ 0.1 & \textbf{58.0} $\pm$ 0.1 \\
        E2CNN + PE (train and test) & 79.6 $\pm$ 1.2 & 58.5 $\pm$ 0.7 \\
        E2CNN + PE (train only; ours) & 79.8 $\pm$ 0.8 & 58.2 $\pm$ 0.7 \\
        \bottomrule
      \end{tabular}
      \end{sc}
    \end{small}
  \end{center}
  \vskip -0.1in
\end{table}

\vskip -0.05 in
\subsection{Steerable E(3)-Equivariant GNNs}
\vskip -0.05 in
Finally, we evaluate our method on the charged-particle N-Body forecasting task \cite{kipf_neural_2018} using SEGNN \cite{brandstetter_geometric_2022}.
We use 3000 training samples and 2000/2000 validation/test samples, following the data split of \cite{pertigkiozoglou_improving_2025}, and train for 2000 epochs.
We select the checkpoint with the lowest validation loss and report the corresponding test MSE on final particle positions averaged over 5 random seeds.
We first tune the optimizer hyperparameters (learning rate and weight decay) for the equivariant SEGNN baseline and then fix them for all methods. 
We then tune only the relaxation-specific hyperparameters for each relaxed variant using the same validation criterion (see additional details in \Cref{sec:app-segnn}). \\
The results in \Cref{tab:segnn-nbody} show that both relaxation methods improve over the equivariant baseline on this task. 
Our PE relaxation achieves an average test MSE of $41.9$, performing better than the dense-layer CR method ($43.2$) and improving over the baseline ($44.3$).
This confirms that PE can capture the optimization benefits of constraint relaxation as effectively as dense layers, with fewer parameters (147K for the Baseline and PE vs.~344K for CR). 
We note that due to the small scale of this model, we have not observed any significant change in training time.
\blue{Additional results in \Cref{sec:app-segnn-ace} show that PE can also replace the non-equivariant MLP relaxation inside the ACE \cite{manolache_learning_2025} constrained-optimization framework.}

\begin{table}[t]
  \caption{\textbf{N-Body (charged particles) forecasting with SEGNN.}
  We report test MSE on final particle positions evaluated at the checkpoint with the lowest validation loss (mean $\pm$ std over 5 seeds). }
  \label{tab:segnn-nbody}
  \begin{center}
    \begin{small}
      \begin{sc}
      \begin{tabular}{lc}
        \toprule
    Model & Test MSE ($\times 10^{-4}$) $\downarrow$ \\
    \midrule
    SEGNN \cite{brandstetter_geometric_2022}        & 44.3 $\pm$ 1.4  \\
    SEGNN-CR \cite{pertigkiozoglou_improving_2025} & 43.2 $\pm$ 1.9  \\             
    SEGNN-PE (Ours)                               & 41.9 $\pm$ 1.5 \\ 
    \bottomrule
  \end{tabular}
      \end{sc}
    \end{small}
  \end{center}
  \vskip -0.1in
\end{table}
\vskip -0.05 in
\section{Discussion}
\vskip -0.05 in
In this paper, we introduce a novel framework for equivariance constraint relaxation based on annealing Positional Encodings (PE).
Our approach is based on the insight that the optimization difficulties of strictly equivariant models can be mitigated by temporarily breaking symmetry, allowing the model to traverse a less constrained, generally non-equivariant hypothesis class during training.
Unlike previous methods that rely on computationally expensive dense layers, our framework achieves this via a simple symmetry-breaking signal (PE) that introduces negligible overhead.

\textbf{Efficiency and scalability.} 
A primary contribution of our work is improving the scalability of constraint relaxation. 
While dense-layer relaxation methods incur a quadratic growth in parameters and training time with respect to model width, our PE-based method maintains practically the same computational cost of the equivariant baseline.
For instance, on the VN-PointNet architecture, the dense-layer method (CR) increased the parameter count from 2M to roughly 9M and the training time from 28s to 98s per epoch (\Cref{tab:pointcloud-results}).
In contrast, our PE relaxation incurred negligible overhead in parameters or time while achieving superior accuracy ($79.6\%$ vs $74.0\%$). 
This efficiency is critical for applying relaxation techniques to large-scale architectures, or when considering equivariant models with sparse weight sharing schemes relative to the input dimension \cite{navon_equivariant_2024}.

\textbf{Symmetry misalignment and optimization.} 
Real-world data often satisfy an assumed symmetry only approximately (e.g., due to boundary effects in images \cite{kayhan_translation_2020, islam_how_2020}, boundary conditions in physical systems \cite{wang_approximately_2022}, and measurement artifacts \cite{hu_point_2022}), in which case enforcing exact constraints may degrade performance \cite{wang_approximately_2022}.
Our experiments on 2D grids illustrate the optimization challenges that may arise in such cases.
While AF-XCiT is designed for fractional \emph{cyclic} translation invariance, natural images do not strictly adhere to this group structure (e.g., due to non-cyclic boundary effects), and the smoke plume dynamics only approximately satisfy the $C_4$ symmetry of E2CNN.
This mismatch may explain why retaining PE at test time improves over the equivariant baselines (\Cref{tab:af-xcit,tab:smoke}). Notably, annealing PE to zero still retains most of this gain, e.g., $0.5$ of the $0.8$ accuracy gain on AF-XCiT.
This suggests that even when strict invariance is enforced at inference, the optimization dynamics benefit from a relaxed trajectory that temporarily accommodates the approximate nature of the data symmetries.
\blue{This is further supported by the training-dynamics plots in \Cref{sec:app-training-dynamics}.}

\textbf{Limitations and future work.}
Despite the demonstrated advantages of our method, we acknowledge specific limitations of our work.
First, our method requires selecting a few configurations such as the PE type and schedule, which are architecture-dependent choices; we propose practical defaults in \Cref{app:recommendations}.
Second, our theory motivates rather than guarantees the method: \cref{prop:universality-via-pe} covers only concatenated PE with a trivial-stabilizer anchor on a universal equivariant backbone, and does not directly apply to our additive or Fixed-PE configurations.
Third, the accuracy gains vary between the evaluated architectures, and are more modest on already strong baselines, e.g.~VN-DGCNN versus VN-PointNet on ModelNet.
Fourth, we inject PE only at the input; which layers benefit most from symmetry breaking remains open (\cref{sec:app-segnn-ace,sec:app-pe-placement}).
Finally, while our evaluation focused on rotations and translations, applying PE relaxation to other groups, such as permutations, remains an open question, and could be especially useful in domains where dense-layer relaxation is computationally infeasible.

More broadly, the principle of relaxing constraints via PE has broader implications beyond geometric deep learning.
\blue{Related work} in large language models (LLMs) has shown that removing positional embeddings at inference improves length generalization \cite{gelberg_extending_2025}.
This phenomenon parallels our findings: PE provides useful signals that improve training but can be dropped at test time.
Future work could investigate this intersection further, for example, leveraging our scheduling and regularization in LLM training.

\begin{ack}

The research of DS was funded by the European Union (ERC, A-B-C-Deep, 101039436). Views and opinions expressed are however those of the author only and do not necessarily reflect those of the European Union or the European Research Council Executive Agency (ERCEA). Neither the European Union nor the granting authority can be held responsible for them. The research of DS was also supported by the Council for Higher Education in Israel under the Moonshot Project.  The research is also supported by the NVIDIA Academic Grant Program using donated A100 GPU-Hours on Brev.
HM
is supported by the Israel Science Foundation through a personal grant (ISF 264/23)
and an equipment grant (ISF 532/23), and by the Career Advancement Chairs in Artificial
Intelligence – Schmidt Futures.
\end{ack}

\bibliographystyle{unsrtnat}

\bibliography{references}

\newpage
\appendix
\section{Formal Proofs}
\subsection{A Formal Proof of Proposition 4.2} \label{sec:equivariance-error-bound-proof}

\textbf{Proposition 4.2.}
\textit{
Let $f_\theta = f^L \circ \dots \circ f^1$ be a composition of $L$ layers, where each layer $f^l$ is $M$-Lipschitz continuous as a map between its corresponding input and output representation spaces. 
    Assume that the group action is isometric.
    The equivariance error of the relaxed model $h_{\lambda}(x) = f_\theta \left( x, \lambda p_0 \right)$ is bounded by:
    \begin{equation*}
    \lVert h_{\lambda} \left( g \cdot x \right) - g \cdot h_{\lambda}\left( x \right) \rVert \leq  2 \lambda M^L \lVert p_0 \rVert
    \end{equation*}
}
\begin{proof}
We seek to bound the equivariance error defined as $\Delta\left(x, g\right) = \left\lVert h_\lambda\left(g \cdot x\right) - g \cdot h_\lambda\left(x\right) \right\rVert$.
Substituting the definition of the relaxed model $h_\lambda\left(x\right) = f_\theta\left(x, \lambda p_0\right)$:
\begin{equation}
    \Delta\left(x, g\right) = \left\lVert f_\theta\left(g \cdot x, \lambda p_0\right) - g \cdot f_\theta\left(x, \lambda p_0\right) \right\rVert
\end{equation}
Since the underlying network $f_\theta$ is strictly equivariant with respect to the joint action on the input tuple $\left(x, p\right)$, we have the property:
\begin{equation}
    g \cdot f_\theta\left(x, \lambda p_0\right) = f_\theta\left(g \cdot x, g \cdot \left(\lambda p_0\right)\right)
\end{equation}
Substituting this back into the error expression, we see that the equivariance error is equivalent to the stability of the network with respect to a perturbation of the anchor $p_0$:
\begin{equation}
    \Delta\left(x, g\right) = \left\lVert f_\theta\left(g \cdot x, \lambda p_0\right) - f_\theta\left(g \cdot x, \lambda g \cdot p_0\right) \right\rVert
\end{equation}

We next bound $\Delta\left(x, g\right)$ by recursion. 
Let $z_l$ denote the feature activation at layer $l$ for the input $\left( g\cdot x, \lambda p_0 \right)$, and $\tilde{z}_l$ denote the activation for the input $\left( g\cdot x, \lambda g \cdot p_0 \right)$.
Concretely,
\begin{equation}
\begin{aligned}
  & z_0 = \left(g \cdot x, \lambda p_0\right)\,, &z_l = f^l\left(z_{l-1}\right) \\
  & \tilde{z}_0 = \left(g \cdot x, \lambda g \cdot p_0\right), &\tilde{z}_l = f^l\left(\tilde{z}_{l-1}\right) \,.
\end{aligned}    
\end{equation}
 
The term we wish to bound is the distance at the final layer, $\delta_L = \left\lVert z_L - \tilde{z}_L \right\rVert$. We proceed by induction.

\textbf{Base Case ($l=0$):}
The distance at the input level depends only on the positional encodings, as the signal component $g \cdot x$ is identical in both trajectories:
\begin{equation}
    \delta_0 = \left\lVert z_0 - \tilde{z}_0 \right\rVert = \left\lVert \left(g \cdot x, \lambda p_0\right) - \left(g \cdot x, \lambda g \cdot p_0\right) \right\rVert
\end{equation}
\begin{equation}
    \delta_0 = \left\lVert \lambda p_0 - \lambda g \cdot p_0 \right\rVert = \lambda \left\lVert p_0 - g \cdot p_0 \right\rVert
\end{equation}
By the Triangle Inequality
and assuming the group action is isometric, we get
\begin{equation}
    \delta_0 \leq 2 \lambda \left\lVert p_0 \right\rVert \,.
\end{equation}

\textbf{Inductive Step:}
Assume the distance at layer $l-1$ is $\delta_{l-1}$. Since the layer function $f^l$ is $M$-Lipschitz continuous, the distance at layer $l$ is bounded by:
\begin{equation}
    \delta_l = \left\lVert f^l\left(z_{l-1}\right) - f^l\left(\tilde{z}_{l-1}\right) \right\rVert \leq M \left\lVert z_{l-1} - \tilde{z}_{l-1} \right\rVert = M \delta_{l-1} \,.
\end{equation}

\textbf{Recursion:}
Applying the inductive step recursively from the final layer $L$ back to the input:
\begin{equation}
    \delta_L \leq M \delta_{L-1} \leq M \left( M \delta_{L-2} \right) \leq \dots \leq M^L \delta_0 \,.
\end{equation}
Substituting the bound for $\delta_0$ derived in the base case:
\begin{equation}
    \Delta\left(x, g\right) = \delta_L \leq M^L \left( 2 \lambda \left\lVert p_0 \right\rVert \right) = 2 \lambda M^L \left\lVert p_0 \right\rVert \,.
\end{equation}
\end{proof}

\section{Ablation study}
\label{sec:app-ablation-vn}
We next ablate the main design choices of our PE relaxation on the Vector Neurons models. 
In both architectures, we examine the impact of the main components of our methodology by
selectively removing them. For the schedule of the symmetry-breaking coefficient $\lambda$, we
compare the best-performing decay schedule for each architecture against a no-scheduled-decay
control, i.e., $\lambda(t)\equiv 1$ throughout training, denoted by \textsc{None}. For the
regularization strength $\beta$, we compare the best-performing nonzero value against no explicit
PE-magnitude regularization, i.e., $\beta=0$. For the PE scale $\gamma$, we compare the full
setting in which $\gamma$ is learned during training (\cmark) against an ablated variant in which
$\gamma$ is kept fixed at its initialization (\xmark). 
The remaining PE design choices are fixed to
the best configurations selected by the relaxation hyperparameter sweep in
\Cref{tab:pointnet_sweep,tab:pe_optimal_params}: additive Sinusoidal-PE for VN-PointNet
(with best decay schedule Linear and best nonzero regularization $\beta=10^{-3}$), and
concat-based Sinusoidal-PE for VN-DGCNN (with best decay schedule Triangular and best
nonzero regularization $\beta=10^{-3}$).
For each configuration, we report in \Cref{tab:ablation-vn-pointnet,tab:ablation-vn-dgcnn} two quantities. 
First, we report the best validation accuracy of the final strict equivariant model evaluated without PE at test time. 
Second, we report the best validation accuracy of the relaxed model evaluated with PE, which is informative when the positional signal is not fully annealed and the resulting model remains only approximately equivariant.

\paragraph{Results.}
The ablations in \Cref{tab:ablation-vn-pointnet,tab:ablation-vn-dgcnn} show two main trends.
First, learning the PE scale $\gamma$ is critical for transferring the benefit of PE relaxation to the
strict equivariant model evaluated without PE. In VN-PointNet, fixing $\gamma$ leads to severe
degradation in both the no-scheduled-decay and linear settings (rows 1,3,5,7), whereas learning
$\gamma$ recovers strong strict-equivariant performance (rows 2,4,6,8). In VN-DGCNN, the same
effect appears more subtly: without a learnable $\gamma$, the no-scheduled-decay control can perform
well only when PE is retained at evaluation, but drops substantially after PE removal (rows 1 and 3).
Learning $\gamma$ largely closes this gap (rows 2 and 4). Second, the role of explicit scheduling is
architecture-dependent. For VN-PointNet, the no-scheduled-decay control with regularization and
learnable $\gamma$ gives the strongest value in this ablation (row 4), suggesting that the learnable
scale and PE-magnitude regularization can already reduce reliance on the positional signal. For
VN-DGCNN, explicit triangular annealing is important for the strict equivariant model, especially
when $\gamma$ is fixed, improving the without-PE result from rows 1/3 to rows 5/7. Importantly,
\textsc{None} is included only as a diagnostic control and is not treated as an annealing schedule in
our main methodology; the main VN results use the decaying schedules reported in
\Cref{tab:pointnet_params}. Overall, these ablations support treating PE as a transient
symmetry-breaking mechanism during training rather than as a permanent architectural modification.

\begin{table*}[h]
\caption{\textbf{Ablation study for VN-PointNet on ModelNet40.}
We vary the annealing schedule, regularization strength, and whether the PE scale is learnable.
Results are mean $\pm$ std over 3 seeds.
Training and evaluation use randomly rotated point clouds ($SO(3)/SO(3)$).
}
  \label{tab:ablation-vn-pointnet}
  \begin{center}
    \begin{small}
      \begin{sc}
        \begin{tabular}{clcccc}
          \toprule
          & Annealing Schedule $\lambda$ & Reg.\ $\beta$ & \doublelinecell[]{PE learnable\\scale $\gamma$} & \doublelinecell[]{Accuracy equivariant\\(w/o\ PE)} & \doublelinecell[]{Accuracy\\(with PE)} \\
          \midrule
          1 & None & 0      & \xmark & 4.5 $\pm$ 0.3  & 35.6 $\pm$ 0.3 \\
          2 & None & 0      & \cmark & 79.7 $\pm$ 2.7 & 79.8 $\pm$ 2.1 \\
          3 & None & 0.001  & \xmark & 4.5 $\pm$ 0.3  & 35.6 $\pm$ 0.3 \\
          4 & None & 0.001  & \cmark & \textbf{80.7 $\pm$ 0.8} & \textbf{80.8 $\pm$ 0.9} \\
          5 & Linear   & 0      & \xmark & 50.9 $\pm$ 0.7 & 51.5 $\pm$ 0.3 \\
          6 & Linear   & 0      & \cmark & 80.4 $\pm$ 1.0 & 80.4 $\pm$ 0.9 \\
          7 & Linear   & 0.001  & \xmark & 50.9 $\pm$ 0.7 & 51.5 $\pm$ 0.3 \\
          8 & Linear   & 0.001  & \cmark & 80.1 $\pm$ 1.3 & 80.2 $\pm$ 1.3 \\
          \bottomrule
        \end{tabular}
      \end{sc}
    \end{small}
  \end{center}
  \vskip -0.1in
\end{table*}

\begin{table*}[h]
\caption{\textbf{Ablation study for VN-DGCNN on ModelNet40.}
We repeat the same ablation over the annealing schedule, regularization strength, and learnable PE scale.
Results are mean $\pm$ std over 3 seeds.
Training and evaluation use randomly rotated point clouds ($SO(3)/SO(3)$).
}

\label{tab:ablation-vn-dgcnn}
  \begin{center}
    \begin{small}
      \begin{sc}
        \begin{tabular}{clcccc}
          \toprule
          & Annealing Schedule $\lambda$ & Reg.\ $\beta$ & \doublelinecell[]{PE learnable\\scale $\gamma$} & \doublelinecell[]{Accuracy equivariant\\(w/o\ PE)} & \doublelinecell[]{Accuracy\\(with PE)} \\
          \midrule
          1 & None   & 0      & \xmark  & 74.1 $\pm$ 3.6 & 88.8 $\pm$ 0.1 \\
          2 & None   & 0      & \cmark  & 88.6 $\pm$ 0.2 & 88.6 $\pm$ 0.1 \\
          3 & None   & 0.001  & \xmark  & 74.1 $\pm$ 3.6 & 88.8 $\pm$ 0.1 \\
          4 & None   & 0.001  & \cmark  & 88.5 $\pm$ 0.4 & 88.5 $\pm$ 0.3 \\
          5 & Triangular & 0      & \xmark  & 88.1 $\pm$ 0.5 & 88.6 $\pm$ 0.4 \\
          6 & Triangular & 0      & \cmark  & 88.2 $\pm$ 0.1 & 88.2 $\pm$ 0.1 \\
          7 & Triangular & 0.001  & \xmark  & 88.5 $\pm$ 0.5 & 88.6 $\pm$ 0.4 \\
          8 & Triangular & 0.001  & \cmark  & 88.5 $\pm$ 0.3 & 88.84$\pm$ 0.3 \\
          \bottomrule
        \end{tabular}
      \end{sc}
    \end{small}
  \end{center}
  \vskip -0.1in
\end{table*}

\FloatBarrier

\subsection{Stabilizer of Fixed-PE}
\label{app:fixed-pe-stabilizer}

We further clarify a subtle point concerning the Fixed-PE variant used in \Cref{sec:experiments}. \Cref{prop:universality-via-pe} assumes that the positional
anchor has a trivial stabilizer. However, the broadcast Fixed-PE anchor used in our
main ablations consists of a single learnable vector \(v\in\mathbb{R}^3\), repeated for
all points:
\[
    p_k = v, \qquad k=1,\ldots,N .
\]
For \(v\neq 0\), its \(SO(3)\) stabilizer is
\[
    \mathrm{Stab}_{SO(3)}(p)
    =
    \{R\in SO(3): Rv=v\}
    \simeq SO(2),
\]
corresponding to rotations around the axis defined by \(v\). Thus, Fixed-PE breaks
almost all rotations, but it does not strictly satisfy the trivial-stabilizer premise of
\Cref{prop:universality-via-pe}. Intuitively, however, this residual stabilizer is a much
smaller subgroup of \(SO(3)\): under a continuous random draw from \(SO(3)\), the
probability that the sampled rotation lies exactly in this \(SO(2)\) stabilizer is zero.
This suggests that, although the distinction matters formally, it is unlikely to have a
large practical effect in the randomized rotation setting used in our experiments.
In particular, since VN inputs are randomly rotated during training and testing (\cref{sec:app-point-cloud}), the data have no consistent $z$-axis with which the broadcast Fixed-PE vector could align.

To check this empirically, we run an additional fixed-frame concat ablation. Instead
of a single broadcast vector, we use three learned axis-aligned vector channels,
\[
    p^{\mathrm{frame}}
    =
    \big(\alpha_1 e_1,\alpha_2 e_2,\alpha_3 e_3\big),
\]
which, after flattening, corresponds to
\[
    (\alpha_1,0,0,\;0,\alpha_2,0,\;0,0,\alpha_3).
\]
For nonzero \(\alpha_1,\alpha_2,\alpha_3\), any \(R\in SO(3)\) that fixes this ordered
frame must satisfy \(R=I\). This FixedSO-PE variant therefore satisfies the
trivial-stabilizer condition while keeping the same qualitative fixed, learnable,
concat-based PE structure. We treat this variant only as a diagnostic ablation and do
not consider it in the rest of the experiments, since it introduces multiple vector
channels and therefore cannot be directly used in the additive PE setting without
changing the input representation.

\Cref{tab:fixed-so-ablation} directly compares the original broadcast Fixed-PE
implementation used in the sweep against the trivial-stabilizer
FixedSO-PE version on both VN-PointNet and VN-DGCNN. 
To isolate
the effect of the stabilizer, we use the best hyperparameters found for concat-based
PE with Fixed-PE in the original sweep and do not re-tune the FixedSO-PE variant.
Following the main VN results, we report validation accuracy only.

\begin{table}[t]
\centering
\caption{
\textbf{Fixed-PE stabilizer ablation on Vector Neurons.}
We compare the original broadcast Fixed-PE anchor, whose
\(SO(3)\) stabilizer is \(SO(2)\), with the FixedSO-PE frame variant, whose \(SO(3)\) stabilizer is trivial. All runs use concat-based
PE and the best hyperparameters found for concat-based Fixed-PE in the original
sweep. Results are validation accuracy, reported as mean \(\pm\) std over 3 seeds.
}
\label{tab:fixed-so-ablation}
\begin{tabular}{llcc}
\toprule
Model &
PE variant &
\(\mathrm{Stab}_{SO(3)}\) &
Val accuracy \\
\midrule
VN-PointNet &
Fixed-PE concat  &
\(SO(2)\) &
\(69.4 \pm 1.3\) \\
VN-PointNet &
FixedSO-PE concat  &
\(\{I\}\) &
\(70.5 \pm 1.6\) \\
\midrule
VN-DGCNN &
Fixed-PE concat  &
\(SO(2)\) &
\(88.4 \pm 0.2\) \\
VN-DGCNN &
FixedSO-PE concat  &
\(\{I\}\) &
\(88.4 \pm 0.3\) \\
\bottomrule
\end{tabular}
\end{table}

As shown in \Cref{tab:fixed-so-ablation}, replacing the broadcast Fixed-PE anchor
with the trivial-stabilizer FixedSO-PE frame does not yield a consistent practical
improvement.
On VN-PointNet, FixedSO-PE gives a small increase over Fixed-PE, while on VN-DGCNN the two variants perform the same.
In both architectures, the FixedSO-PE variant also
does not outperform the corresponding sinusoidal PE configuration used in our main
VN results. These findings support the intuition above: although the original Fixed-PE
does not fully remove the stabilizer in the formal sense required by
\Cref{prop:universality-via-pe}, its residual \(SO(2)\) stabilizer is sufficiently small
that the discrepancy does not appear to have a meaningful practical impact. In
contrast, PE anchors whose 3D vectors span a subspace of dimension at least two,
such as the sinusoidal concat PE used in the main VN experiments, have trivial
\(SO(3)\) stabilizer and therefore satisfy the proposition's no-self-symmetry
condition.

\subsection{Practical recommendations}
\label{app:recommendations}
Our method requires selecting a few configurations, such as the PE type, regularization and decay schedule.
While our experiments do not suggest an architecture-independent optimal recipe, our sweeps (\cref{tab:pe_optimal_params,tab:rapidash_optimal_params,tab:xcit_params,tab:smoke_optimal_params,tab:segnn_optimal_params}) and ablations (\cref{tab:ablation-vn-pointnet,tab:ablation-vn-dgcnn}) suggest a useful default configuration as a starting point.
(i) Use a learnable $\gamma$, initialized to 1. It is the most important component and is needed for the regularizer to affect deterministic encodings.
(ii) Start with Sinusoidal-PE and a linear decay schedule, each of which was selected in most of our final configurations; triangular decay was preferred for concatenative VN-DGCNN and SEGNN, and Fixed-PE for additive VN-DGCNN.
(iii) Use weak regularization: selected values ranged from $10^{-5}$ (AF-XCiT) to $5\cdot10^{-2}$ (SEGNN), and in the VN ablations $\beta$ had little effect once $\gamma$ was learned.
(iv) Prefer additive PE when the input and the PE share a representation.
Concatenative PE matches the setting of \cref{prop:universality-via-pe} but changes the input representation, which can be inconvenient for architectures with fixed feature dimensions or irrep layouts.

\section{Additional Results}
\subsection{Additional scalability analysis for Vector Neurons}
\label{app:vn_scalability}
To further characterize the computational cost gaps in \Cref{tab:pointcloud-results}, we analyze the scaling behavior of the relaxation methods on the VN-PointNet architecture by varying both the model width and the input point-cloud size. 
For each method, we use the optimal configuration selected in the main experiments and report averages over 3 random seeds.

\subsubsection{Model width and input-size scaling}
\begin{figure*}[h]
  \centering
  \begin{subfigure}[b]{0.31\textwidth}
    \includegraphics[width=\linewidth]{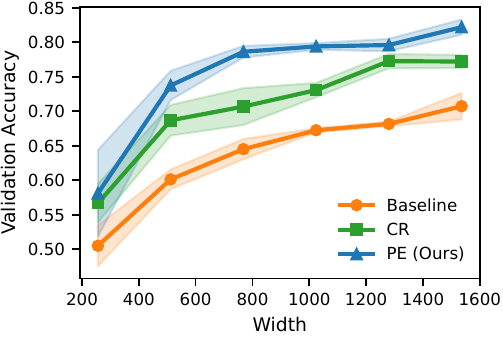}
    \caption{Validation Accuracy}
  \end{subfigure}
  \hfill
  \begin{subfigure}[b]{0.31\textwidth}
    \includegraphics[width=\linewidth]{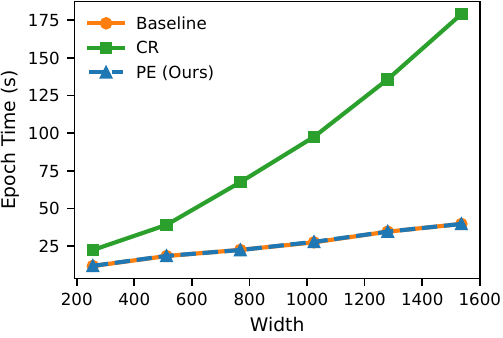}
    \caption{Training Time}
  \end{subfigure}
  \hfill
  \begin{subfigure}[b]{0.31\textwidth}
    \includegraphics[width=\linewidth]{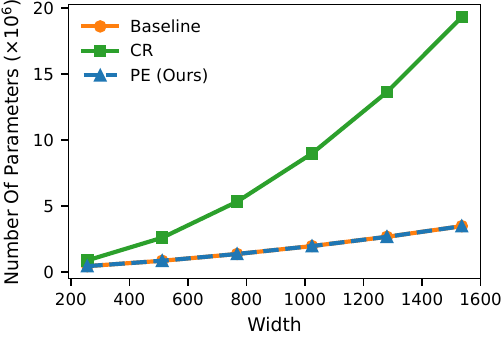}
    \caption{Trainable Parameter Count}
  \end{subfigure}

  \caption{
  \textbf{Additional model-width scaling results on VN-PointNet (ModelNet40).}
  We vary model width by multiplying all channel dimensions by the same factor and plot results against the maximum channel dimension (x-axis).
  (a) Best validation accuracy (mean $\pm$ std over 3 seeds).
  (b) Training time per epoch on a single RTX A6000.
  (c) Trainable parameters during training.
  PE consistently improves the equivariant baseline in accuracy while tracking its training cost, whereas CR's gains come with rapidly increasing overhead at larger widths.
  }
  \label{fig:vn-pointnet-width-app}
\end{figure*}

\paragraph{Model width.}
We scale the model width by multiplying all channel sizes by a constant factor, and use the width of the widest layer as the x-axis in \Cref{fig:vn-pointnet-width-app}. \Cref{fig:vn-pointnet-width-app}a shows that all methods follow a similar accuracy trend as width increases, with PE consistently matching or exceeding both the strict baseline and CR. More importantly, \Cref{fig:vn-pointnet-width-app}b,c further confirm the main scalability claim: PE tracks the strict baseline almost exactly in both training time and parameter count across widths, whereas the overhead of CR grows much more rapidly. As discussed in more detail below, this growth is consistent with the quadratic width dependence induced by the additional dense relaxation branch.

\FloatBarrier

\begin{figure}[h]
  \centering
  \begin{subfigure}[b]{0.31\textwidth}
    \includegraphics[width=\linewidth]{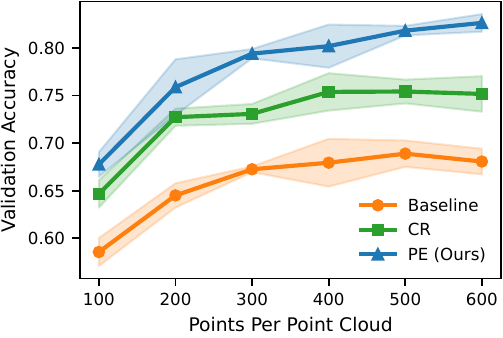}
    \caption{Validation Accuracy}
  \end{subfigure}
  \begin{subfigure}[b]{0.31\textwidth}
    \includegraphics[width=\linewidth]{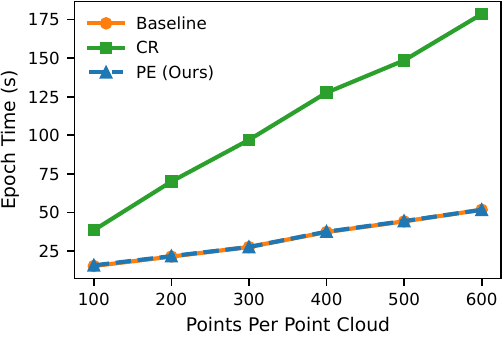}
    \caption{Training Time}
  \end{subfigure}
  \caption{
  \textbf{Additional input-size scaling results on VN-PointNet (ModelNet40).}
  We vary the number of points per point cloud while keeping the architecture fixed.
  (a) Best validation accuracy (mean $\pm$ std over 3 seeds).
  (b) Training time per epoch on a single RTX A6000.
  All methods exhibit approximately linear scaling in epoch time with the number of points, while PE consistently exceeds the accuracy of CR at similar cost to the equivariant baseline.
  }
  \label{fig:vn-pointnet-pc-size-app}
\end{figure}

\paragraph{Point-cloud size.}
We next vary the number of points per point cloud while keeping the network architecture fixed. \Cref{fig:vn-pointnet-pc-size-app}a shows that PE outperforms the other methods across all tested input sizes while preserving the same overall scaling trend as the strict baseline. In contrast, CR incurs a substantially higher per-epoch cost across the entire range. \Cref{fig:vn-pointnet-pc-size-app}b shows that the training time of all methods grows approximately linearly with the number of points, as expected since each method performs roughly fixed computation per point.

\FloatBarrier

\subsubsection{Why CR scales quadratically with width}
An equivariant linear layer in the Vector Neurons framework mixes vector channels using scalar weights:
\begin{equation}
f_{\mathrm{eq},W}(X) = WX,\qquad W \in \mathbb{R}^{c_{\mathrm{out}}\times c_{\mathrm{in}}},\quad X \in \mathbb{R}^{c_{\mathrm{in}}\times 3},
\end{equation}
such that the parameter count is $P_{\mathrm{eq}} = c_{\mathrm{out}}c_{\mathrm{in}}$, where $X$ is a representation of a single point in the point cloud and $c_{\mathrm{in}}, c_{\mathrm{out}}$ are the input and output channels.

\begin{figure}[h]
  \centering
  \begin{subfigure}[b]{0.48\linewidth}
    \includegraphics[width=\linewidth]{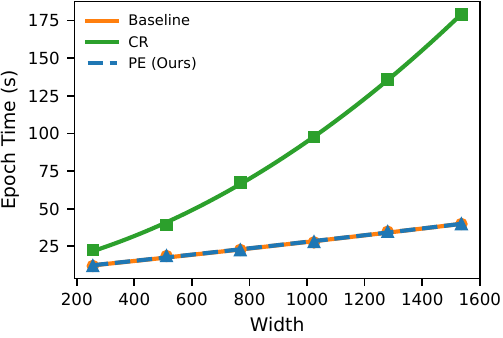}
    \caption{Training time per epoch vs.\ width}
  \end{subfigure}
  \hfill
  \begin{subfigure}[b]{0.48\linewidth}
    \includegraphics[width=\linewidth]{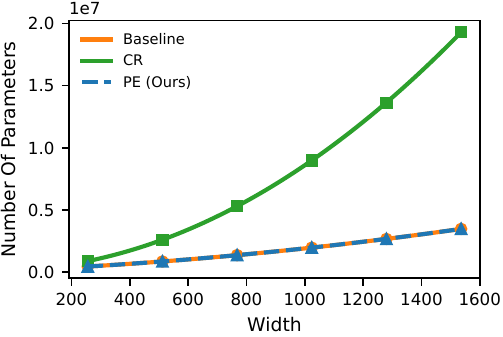}
    \caption{Trainable parameters vs.\ width}
  \end{subfigure}
  \caption{
  \textbf{Quadratic fits for VN-PointNet width scaling.}
  We report least-squares quadratic fits to epoch time (left) and trainable parameters (right) for the strictly equivariant baseline, dense-layer constraint relaxation (CR), and positional-encoding relaxation (PE).
  }
  \label{fig:vn-pointnet-quadratic-fit}
\end{figure}

In the relaxed CR variant, an unconstrained linear map is added on top of the flattened coordinates of the vector channels. Writing $\mathrm{vec}(\cdot)$ for flattening and $\mathrm{uvec}(\cdot)$ for the inverse reshape,
\begin{equation}
f_{\mathrm{relaxed}}(X) = WX + \lambda \,\mathrm{uvec}\!\left(W_{\mathrm{neq}}\,\mathrm{vec}(X)\right), \qquad
W_{\mathrm{neq}} \in \mathbb{R}^{(3c_{\mathrm{out}})\times(3c_{\mathrm{in}})}.
\end{equation}
The dense term has $(3c_{\mathrm{out}})(3c_{\mathrm{in}})=9c_{\mathrm{out}}c_{\mathrm{in}}$ parameters, hence the leading-order parameter count becomes
\begin{equation}
P_{\mathrm{relaxed}}
\approx
\underbrace{c_{\mathrm{out}}c_{\mathrm{in}}}_{\text{strict}}
+
\underbrace{9c_{\mathrm{out}}c_{\mathrm{in}}}_{\text{dense}}
=
10\,c_{\mathrm{out}}c_{\mathrm{in}}.
\end{equation}
This implies that parameter count and training time scale quadratically with width in both the baseline and CR models, while the \emph{overhead} introduced by CR also grows quadratically. We empirically verify this by fitting quadratic curves to the width-scaling results in \Cref{fig:vn-pointnet-quadratic-fit}.

Concretely, fitting $T(W)=aW^2+bW+c$ to epoch time yields
\begin{align*}
T_{\mathrm{Baseline}}(W) &= 1.22 \times 10^{-6} W^2 + 0.0194 W + 7.29 \qquad (R^2 = 0.9958),\\
T_{\mathrm{CR}}(W) &= 4.70 \times 10^{-5} W^2 + 0.0388 W + 8.76 \qquad (R^2 = 0.9997),\\
T_{\mathrm{PE}}(W) &= 1.12 \times 10^{-6} W^2 + 0.0195 W + 7.33 \qquad (R^2 = 0.9962).
\end{align*}
Similarly, fitting $P(W)=aW^2+bW+c$ to the trainable parameter counts yields
\begin{align*}
P_{\mathrm{Baseline}}(W) &= 0.750 W^2 + 1018.17 W + 144324.10 \qquad (R^2 = 1.0000),\\
P_{\mathrm{CR}}(W) &= 7.49 W^2 + 947.21 W + 158714.20 \qquad (R^2 = 1.0000),\\
P_{\mathrm{PE}}(W) &= 0.750 W^2 + 1018.17 W + 144325.10 \qquad (R^2 = 1.0000).
\end{align*}
The fitted quadratic coefficients satisfy $7.49/0.750 \approx 9.99$, matching the expected $10\times$ factor above. This agreement reflects that the relaxed dense map operates on a width-$3c_{\mathrm{in}}$ input and produces a width-$3c_{\mathrm{out}}$ output, contributing $9c_{\mathrm{out}}c_{\mathrm{in}}$ parameters, and corresponding compute, on top of the strict $c_{\mathrm{out}}c_{\mathrm{in}}$ channel mixing.

\FloatBarrier

\paragraph{Run time measurements.}
The reported training times in Table~\ref{tab:pointcloud-results} and \Cref{fig:vn-pointnet-width-app,fig:vn-pointnet-pc-size-app,fig:vn-pointnet-quadratic-fit} represent the total wall-clock time divided by the number of epochs. This includes periodic validation and any logging or checkpointing overhead, and therefore slightly overestimates the pure training cost. We observe a larger relative increase in training time for CR on VN-PointNet and VN-DGCNN compared to \cite{pertigkiozoglou_improving_2025}. However, while the absolute CR training times are relatively close between the two works (e.g., $97.6$s vs.~$80$s for VN-PointNet), our baseline training time is significantly lower ($27.6$s vs.~$75$s). We believe this suggests that the baseline measurements in the prior work were subject to larger fixed overheads, such as data loading or system latency, which dominated the epoch time. In our setup, where such overheads appear minimized, the computational cost of the relaxation layers constitutes a larger fraction of total runtime, resulting in a higher relative increase. This interpretation is further supported by the disparity in VN-DGCNN runtime ratios observed between \cite{pertigkiozoglou_improving_2025} and \cite{manolache_learning_2025}, indicating that relative training costs are sensitive to hardware configuration and optimization regime.

\subsection{Training dynamics}
\label{sec:app-training-dynamics}
\begin{figure*}[h]
  \centering
  \begin{subfigure}[b]{0.48\textwidth}
    \includegraphics[width=\linewidth]{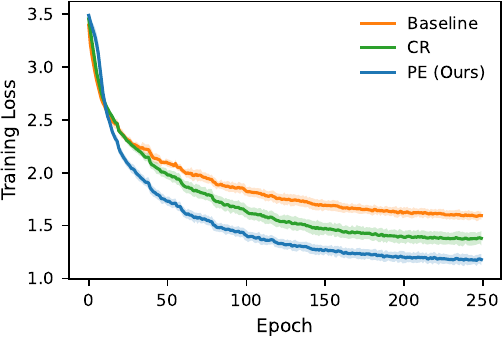}
    \caption{Training Loss}
  \end{subfigure}
  \hfill
  \begin{subfigure}[b]{0.48\textwidth}
    \includegraphics[width=\linewidth]{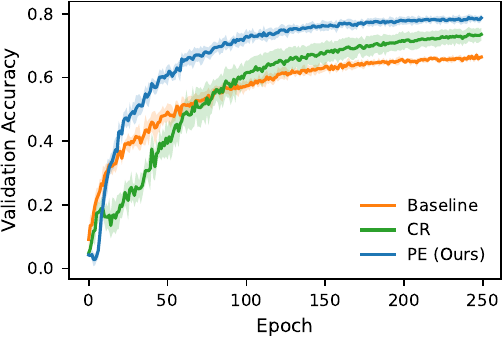}
    \caption{Validation Accuracy}
  \end{subfigure}

  \caption{
  \textbf{Training dynamics on VN-PointNet (ModelNet40).}
  We compare the strict equivariant baseline, dense-layer constraint relaxation (CR), and positional-encoding relaxation (PE).
  (a) Training loss over epochs.
  (b) Validation accuracy over epochs.
  Validation accuracy is evaluated with PE removed.
  PE descends faster and reaches lower training loss through most of training, while also achieving higher validation accuracy after the initial training phase.
  This suggests that the improvement of PE over both the strict baseline and CR is associated with more favorable optimization dynamics.
  }

  \label{fig:pointnet-training-dynamics}
\end{figure*}

\begin{figure*}[h]
  \centering
  \begin{subfigure}[b]{0.48\textwidth}
    \includegraphics[width=\linewidth]{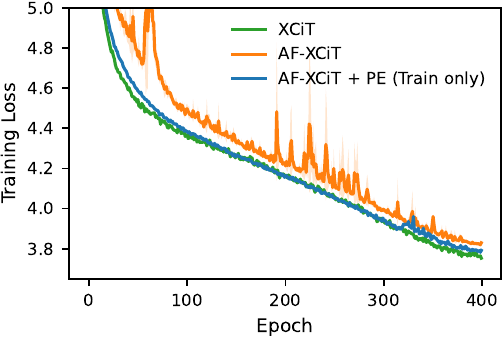}
    \caption{Training Loss}
  \end{subfigure}
  \hfill
  \begin{subfigure}[b]{0.48\textwidth}
    \includegraphics[width=\linewidth]{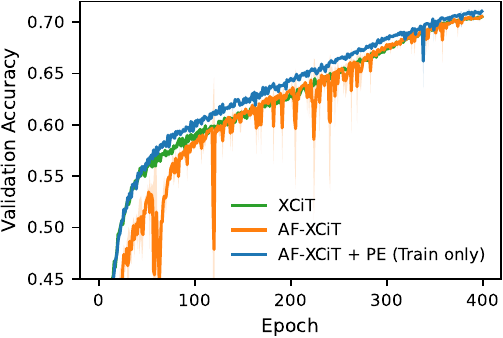}
    \caption{Validation Accuracy}
  \end{subfigure}
  
  \caption{
  \textbf{Training dynamics on ImageNet for XCiT and AF-XCiT.}
  We compare standard XCiT, strict AF-XCiT, and AF-XCiT with train-time PE that is annealed to zero before inference.
  (a) Training loss over epochs.
  (b) Validation accuracy over epochs.
  Validation accuracy is evaluated with PE removed.
  The strict AF-XCiT model exhibits a noisier optimization trajectory, whereas AF-XCiT + PE produces smoother training curves and finishes with slightly improved validation accuracy.
  The PE trajectory remains close to that of standard XCiT while still recovering the original alias-free architecture at test time.
  }
  
  \label{fig:xcit-training-dynamics}
\end{figure*}

We further examine whether the gains of PE are reflected in the optimization trajectory. 
To this end, \Cref{fig:pointnet-training-dynamics,fig:xcit-training-dynamics} report the full training-loss and validation-accuracy curves for representative point-cloud and image-classification architectures.

For VN-PointNet on ModelNet40, the effect is pronounced. 
As shown in \Cref{fig:pointnet-training-dynamics}a, PE reduces the training loss faster than both the strict equivariant baseline and the dense-layer constraint-relaxation baseline (CR), and maintains a lower loss through most of training. 
The validation curves in \Cref{fig:pointnet-training-dynamics}b show a corresponding generalization benefit: after the initial training phase, PE consistently attains higher validation accuracy than both alternatives. 
These curves indicate that PE does not merely improve the final checkpoint, but changes the training trajectory in a way that makes optimization easier.

For ImageNet training with AF-XCiT, the effect is more subtle but still visible. 
In \Cref{fig:xcit-training-dynamics}a, the strict Alias-Free model exhibits a noisier training-loss trajectory, whereas AF-XCiT with train-time PE follows a smoother curve that remains closer to standard XCiT. 
The validation-accuracy curves in \Cref{fig:xcit-training-dynamics}b show that train-time PE also yields a small but consistent improvement over strict AF-XCiT near the end of training. 
Importantly, this improvement is obtained while annealing the PE coefficient to zero before evaluation, so the final model recovers the original alias-free architecture at inference.

Overall, the training curves in \Cref{fig:pointnet-training-dynamics,fig:xcit-training-dynamics} support the central role of PE as a transient optimization aid: temporarily breaking the symmetry improves the path taken during training, while the deployed model can still recover the desired strict equivariant or invariant structure.

\FloatBarrier

\subsection{Adversarial robustness}
\label{sec:app-adversarial-robustness}

A central motivation for equivariant and invariant architectures is robustness to transformations in the modeled symmetry group. 
Here, we compare three inference regimes: strictly equivariant or invariant models, our train-time constraint-relaxation approach that removes PE at inference, and approximate-equivariant variants that keep the same symmetry-breaking PE at test time. 
This distinction isolates whether the optimization benefits of relaxation can be retained without deploying a model that relies on a non-equivariant positional signal. 
Across both translation-invariant image classification and rotation-equivariant point-cloud classification, we find that retaining PE at inference can reduce robustness to adversarial transformations aligned with the modeled symmetry, whereas removing PE at test time recovers the transformation consistency of the strict architecture.

\subsubsection{AF-XCiT}
\label{sec:app-adversarial-robustness-afxcit}

We first evaluate AF-XCiT \cite{michaeli_alias-free_2025}, where fractional shift invariance is introduced specifically to reduce sensitivity to adversarial translations. 
We consider three adversarial shift attacks: cyclic shifts, crop-shift, and bilinear fractional shifts. 
Only the cyclic-shift attack exactly matches the symmetry group modeled by AF-XCiT. 
The crop-shift and bilinear-shift attacks correspond more closely to practical image translations, but they introduce boundary, cropping, or interpolation effects, and therefore are not exact elements of the cyclic translation group used by the alias-free construction.

\begin{figure}[h]
  \centering
  \includegraphics[width=0.48\linewidth]{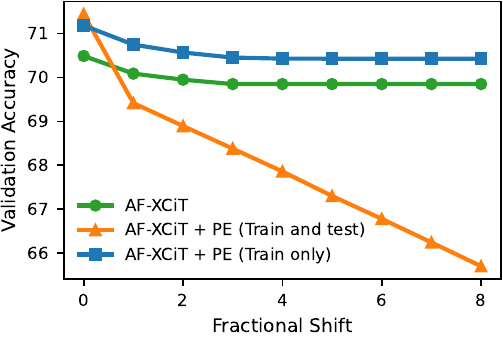}
  \caption{
  \textbf{Adversarial cyclic-shift robustness on ImageNet for AF-XCiT.}
  We report validation accuracy under adversarial cyclic translations, which correspond to the symmetry group modeled by AF-XCiT.
  AF-XCiT + PE (train and test) keeps positional encoding at inference and is therefore only approximately invariant, leading to a substantially larger degradation under this group-aligned attack.
  AF-XCiT + PE (train only) anneals the positional encoding to zero and recovers the original invariant architecture, closely matching the robustness of strict AF-XCiT.
  }
  \label{fig:xcit-adversarial-cyclic}
\end{figure}

As shown in \Cref{fig:xcit-adversarial-cyclic}, the model that keeps PE at test time is substantially more sensitive to adversarial cyclic shifts than the strict invariant variants. 
This confirms that retaining PE at inference breaks the intended group invariance and can reduce robustness precisely in the regime where AF-XCiT is designed to be robust. 
In contrast, AF-XCiT + PE (train only) closely tracks the strict AF-XCiT baseline, showing that annealing PE to zero preserves robustness to the modeled symmetry group while retaining the optimization benefit of train-time relaxation.

\begin{figure}[h]
  \centering
  \begin{subfigure}[b]{0.48\textwidth}
    \includegraphics[width=\linewidth]{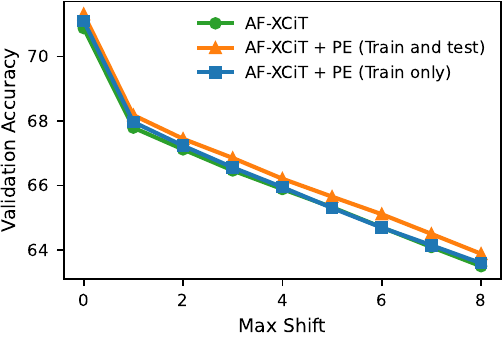}
    \caption{Crop-shift}
  \end{subfigure}
  \hfill
  \begin{subfigure}[b]{0.48\textwidth}
    \includegraphics[width=\linewidth]{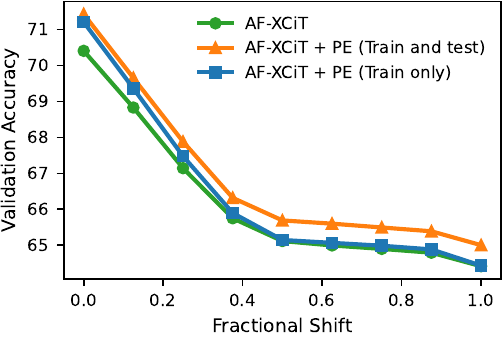}
    \caption{Bilinear fractional shifts}
  \end{subfigure}
  \caption{
  \textbf{Adversarial practical-translation robustness on ImageNet for AF-XCiT.}
  We report validation accuracy under crop-shift and bilinear-shift attacks.
  These transformations reflect practical translations but are not exact elements of the cyclic translation group modeled by AF-XCiT because they introduce boundary, cropping, or interpolation effects.
  Accordingly, the strict invariant and relaxed variants behave more similarly than in the cyclic-shift setting, suggesting that exact cyclic-shift invariance primarily protects against transformations aligned with the modeled symmetry group.
  }
  \label{fig:xcit-adversarial-practical-shifts}
\end{figure}

For the two practical translation attacks, shown in \Cref{fig:xcit-adversarial-practical-shifts}, the gap between the strict and relaxed variants is much smaller. 
This is consistent with the fact that crop-shift and bilinear-shift are off-group transformations: strict cyclic-shift invariance alone does not guarantee robustness to boundary effects, cropping, or interpolation artifacts. 
Thus, the robustness benefit of recovering the strict AF-XCiT architecture is most visible under the group-aligned cyclic-shift attack, while all variants behave more similarly under practical translations outside the formal symmetry group.

\FloatBarrier

\subsubsection{VN-PointNet}
\label{sec:app-adversarial-robustness-vnpointnet}

We next evaluate whether the same principle holds for rotation-equivariant point-cloud classification. 
Using the VN-PointNet + PE configuration from the ablation study with a constant mix coefficient, regularization $\beta=0.001$, and learnable PE scale $\gamma$, we compare two evaluation modes of the same relaxed model. 
The first keeps PE at test time, yielding a relaxed model that can exploit a non-equivariant positional signal. 
The second removes PE at test time, yielding the projected equivariant model that recovers the original strict architecture.

\begin{figure}[h]
  \centering
  \includegraphics[width=0.48\linewidth]{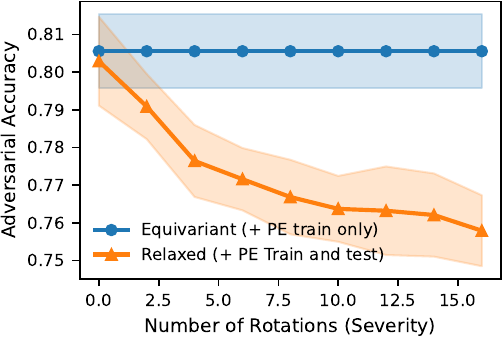}
  \caption{
  \textbf{Adversarial rotation robustness on VN-PointNet (ModelNet40).}
  We compare the relaxed VN-PointNet + PE model evaluated with PE at test time against the projected equivariant model obtained by removing PE at test time.
  Attack severity corresponds to the number of random rotations available to the adversary.
  The projected equivariant model is consistently more robust under adversarial rotations, showing that recovering strict equivariance at inference can improve robustness even when the relaxed model is competitive under standard validation.
  }
  \label{fig:pointnet-adversarial-rotations}
\end{figure}

The adversary samples an increasing number of random rotations and evaluates the model on the worst case among them. 
As shown in \Cref{fig:pointnet-adversarial-rotations}, the projected equivariant model becomes more robust as attack severity increases, even though the relaxed model can be competitive, or slightly better, under standard evaluation. 
This demonstrates the practical value of recovering strict equivariance at inference: PE can help optimization during training without forcing the deployed model to rely on a non-equivariant positional signal that is vulnerable under adversarial rotations.

\FloatBarrier

\subsection{SEGNN with PE under ACE constrained optimization}
\label{sec:app-segnn-ace}

Our main SEGNN experiments follow the train-time constraint-relaxation setting of
\citet{pertigkiozoglou_improving_2025}, where a non-equivariant relaxation branch is
introduced during training and annealed away to recover a strict equivariant model at
inference. We additionally evaluate PE relaxation within the ACE framework of
\citet{manolache_learning_2025}. While CR uses a prescribed relaxation schedule and
regularization to progressively suppress the non-equivariant branch, ACE formulates
the relaxation as a constrained optimization problem and updates the relaxation level
through its optimization procedure.

In this experiment, we keep the ACE/SEGNN training protocol and optimizer unchanged,
but replace each non-equivariant MLP relaxation component used by ACE with a learnable
PE layer. Unlike the input-only PE used throughout the main paper, PE is therefore injected
additively at every intermediate layer of SEGNN. The PE layers serve as the symmetry-breaking
mechanism inside the ACE framework, and we evaluate the recovered strict SEGNN at test time.
For comparison, we also evaluate input-only PE within the same ACE protocol. We also include
an ACE-Linear variant, in which the MLP relaxation is replaced by a single linear layer,
to separate the effect of the ACE optimization framework from the expressivity of the
relaxation branch and to align more closely with CR-style linear relaxations.

The results in \Cref{tab:segnn-ace-pe} show that PE can also serve as an effective
symmetry-breaking mechanism inside a constrained-optimization relaxation framework.
SEGNN with per-layer PE improves over the strict SEGNN baseline and over the ACE-Linear variant,
suggesting that the gain is not only due to replacing the original MLP relaxation with
a lower-capacity branch. At the same time, the original ACE-MLP result remains slightly
stronger, indicating that a more expressive non-equivariant relaxation can still be
beneficial in this small-scale SEGNN setting. In contrast, input-only PE does not improve
over the baseline under ACE, indicating that intermediate placement can be beneficial
in this setting. Overall, these results support the view
that PE is a lightweight and broadly compatible relaxation mechanism: it can be used
either with the simple scheduled relaxation used in the main paper or as the
symmetry-breaking component inside ACE.

\begin{table*}[h]
  \caption{
  \textbf{N-body forecasting with SEGNN under ACE optimization.}
  We report test MSE on final particle positions (lower is better), following the ACE reporting convention. Results for our runs are mean $\pm$ std over 5 seeds.
  Replacing the ACE MLP relaxation with a learnable per-layer PE improves over the strict SEGNN baseline and ACE-Linear, while remaining slightly behind the original ACE-MLP relaxation.
  Input-only PE does not benefit from ACE optimization.
  }
  \label{tab:segnn-ace-pe}
  \begin{center}
    \begin{small}
      \begin{sc}
        \begin{tabular}{lc}
          \toprule
          Method & MSE ($\times 10^{-4}$) $\downarrow$ \\
          \midrule
          Linear \cite{satorras_en_2021} & 819 \\
          SE(3)-Tr.\ \cite{fuchs_se3-transformers_2020} & 244 \\
          RF \cite{kohler_equivariant_2019} & 104 \\
          ClofNet \cite{du_se3_2022} & 65 \\
          EGNN \cite{satorras_en_2021} & 71 \\
          EGNO \cite{xu_equivariant_2024} & 54 \\
          PONITA \cite{bekkers_fast_2024} & 43.0 $\pm$ 1.0 \\
          \midrule
          SEGNN (Baseline) \cite{brandstetter_geometric_2022} & 41.3 $\pm$ 1.0 \\
          SEGNN ACE (MLP) \cite{manolache_learning_2025} & 38.0 $\pm$ 2.0 \\
          SEGNN ACE (Linear) & 40.5 $\pm$ 1.0 \\
          SEGNN PE, input only (ours) & 42.1 $\pm$ 2.0 \\
          SEGNN PE, per layer (ours) & 39.5 $\pm$ 2.0 \\
          \bottomrule
        \end{tabular}
      \end{sc}
    \end{small}
  \end{center}
  \vskip -0.1in
\end{table*}

\FloatBarrier

\subsection{PE placement in Vector Neurons}
\label{sec:app-pe-placement}

Throughout the main paper, PE is injected only at the input of the network, which requires
a single minimal modification to the architecture. Although the subsequent layers remain
equivariant maps, the positional reference propagates through their activations, so the
complete network is non-equivariant while PE is present. To examine whether injecting PE
at additional layers is beneficial, we compare input-only PE with PE injected at every
equivariant layer of VN-PointNet, for both the additive and concatenative variants.
All other hyperparameters follow the configuration of \Cref{tab:pointcloud-results}.

The results in \Cref{tab:vn-pe-placement} show that injecting PE at every layer does not
improve over input-only PE for both PE variants. This suggests that, in this setting, a
single input-level symmetry-breaking signal is sufficient. In contrast, in the SEGNN
experiment under ACE (\Cref{sec:app-segnn-ace}), per-layer PE outperforms input-only PE.
Thus, injecting PE at every layer may be beneficial in some cases, but is not uniformly
beneficial, and the optimal placement of PE remains architecture-dependent.

\begin{table}[h]
  \caption{
  \textbf{PE placement in VN-PointNet (ModelNet40).}
  We report accuracy (mean $\pm$ std over 5 seeds) for PE injected only at the input and at every equivariant layer, following the protocol of \Cref{tab:pointcloud-results}.
  Injecting PE at every layer does not improve over input-only PE.
  }
  \label{tab:vn-pe-placement}
  \begin{center}
    \begin{small}
      \begin{sc}
        \begin{tabular}{lcc}
          \toprule
          Integration & Input only & Every layer \\
          \midrule
          Additive PE & \textbf{79.6} $\pm$ 0.4 & 79.3 $\pm$ 1.6 \\
          Concatenative PE & \textbf{79.3} $\pm$ 0.9 & 78.7 $\pm$ 0.9 \\
          \bottomrule
        \end{tabular}
      \end{sc}
    \end{small}
  \end{center}
  \vskip -0.1in
\end{table}

\FloatBarrier

\subsection{Data efficiency}
\label{sec:app-data-efficiency}

A common motivation for equivariant architectures is improved data efficiency, and a
natural concern is that train-time symmetry breaking might compromise this benefit.
To examine this, we train the strict VN-PointNet baseline and our additive and
concatenative PE variants from \Cref{tab:pointcloud-results} on random subsets containing
$100\%$, $75\%$, $50\%$, and $25\%$ of the ModelNet40 training set, keeping all other
hyperparameters unchanged and evaluating on the full validation set.

As shown in \Cref{tab:vn-data-efficiency}, both PE variants outperform the strict baseline
at every tested subset size, indicating that the benefit of PE persists as the amount of
training data decreases. Thus, train-time symmetry breaking does not appear to compromise
data efficiency in this experiment.

\begin{table}[h]
  \caption{
  \textbf{Data efficiency of VN-PointNet with PE (ModelNet40).}
  We report accuracy (mean $\pm$ std over 5 seeds) when training on a fraction of the training set.
  Both PE variants outperform the strict baseline at every training-set size.
  }
  \label{tab:vn-data-efficiency}
  \begin{center}
    \begin{small}
      \begin{sc}
        \begin{tabular}{lcccc}
          \toprule
          Model & 100\% & 75\% & 50\% & 25\% \\
          \midrule
          VN-PointNet (Baseline) & 67.5 $\pm$ 0.5 & 64.5 $\pm$ 0.2 & 62.3 $\pm$ 1.1 & 48.6 $\pm$ 3.4 \\
          VN-PointNet-PE add. (ours) & \textbf{79.6} $\pm$ 0.4 & \textbf{77.2} $\pm$ 1.2 & 68.8 $\pm$ 4.3 & \textbf{55.1} $\pm$ 2.2 \\
          VN-PointNet-PE concat. (ours) & 79.3 $\pm$ 0.9 & 75.5 $\pm$ 0.5 & \textbf{72.0} $\pm$ 1.2 & 53.1 $\pm$ 4.4 \\
          \bottomrule
        \end{tabular}
      \end{sc}
    \end{small}
  \end{center}
  \vskip -0.1in
\end{table}

\FloatBarrier

\section{Experimental Details}
\label{sec:app-experimental-details}

\subsection{Vector Neurons}
\label{sec:app-point-cloud}

We evaluate our Positional Encoding (PE) relaxation method on the ModelNet40 point-cloud classification task \cite{wu_3d_2015}. 
The dataset consists of point clouds from 40 categories.
Following standard protocols, we represent each point cloud by a sample of 300 points \cite{pertigkiozoglou_improving_2025}. 
We compare our method against two baselines: strictly equivariant Vector Neurons (VN) models (VN-PointNet and VN-DGCNN) \cite{deng_vector_2021} and the dense-layer Constraint Relaxation (CR) method \cite{pertigkiozoglou_improving_2025}.

\paragraph{Implementation Details.}
For a fair comparison, we utilize the training hyperparameters described by \citet{pertigkiozoglou_improving_2025}. 
All models are trained on a single NVIDIA RTX A6000 GPU.
The specific hyperparameters utilized for the final models are listed in \Cref{tab:pointnet_params}.

\paragraph{Rotation Regime.} 
We follow the $SO(3)/SO(3)$ protocol used by the baseline methods \cite{deng_vector_2021, pertigkiozoglou_improving_2025}. 
Specifically, during both training and evaluation, input point clouds are randomly rotated using rotation matrices sampled uniformly from the continuous rotation group $SO(3)$. 
This setup ensures that the task requires learning strictly rotation-robust representations, distinguishing our evaluation from gravity-aligned ($z/z$) settings where objects share a canonical orientation.

\paragraph{Hyperparameter Tuning.}
We tune the relaxation-specific hyperparameters (regularization scale, mix coefficient scheduler, and positional encoding type) for our method. Similarly, we run the same search on the hyperparameters of the dense-layer constraint relaxation approach and report the results of the best configuration averaged over 5 random seeds. The search space used for tuning is detailed in \Cref{tab:pointnet_sweep}.
For the dense-layer CR method, we found the optimal configuration to be a triangular scheduler with regularization $\beta=0.01$ for both architectures. The optimal hyperparameters for our PE method are listed in \Cref{tab:pe_optimal_params}.
The Vector Neurons experiments required approximately 630 GPU-hours in total.  

\begin{table}[h]
    \centering
    \caption{General Hyperparameters for Vector Neurons models.}
    \label{tab:pointnet_params}
    \begin{tabular}{lc}
        \toprule
        \textbf{Hyperparameter} & \textbf{Value} \\
        \midrule
        Dataset & ModelNet40 \\
        Input Points & 300 \\
        Batch Size & 32 \\
        Optimizer & Adam (VN-PointNet) / SGD (VN-DGCNN) \\
        Learning Rate & 0.001 (VN-PointNet) / 0.1 (VN-DGCNN) \\
        Weight Decay & $10^{-4}$ \\
        Scheduler & StepLR (VN-PointNet) / Cosine Annealing (VN-DGCNN) \\
        \bottomrule
    \end{tabular}
\end{table}

\begin{table}[h]
    \centering
    \caption{Hyperparameter Search Space for Point Cloud Classification Models.}
    \label{tab:pointnet_sweep}
    \begin{tabular}{lcc}
        \toprule
        \textbf{Hyperparameter} & \textbf{Method} & \textbf{Values Tested} \\
        \midrule
        Positional Encoding Type & PE (Ours) & \{Sinusoidal, Fixed\} \\
        Mix-coef Scheduler & PE / CR & \{Linear, Triangular\} \\
        Regularization $\beta$ & PE / CR & \{0.1, 0.05, 0.01, 0.005, 0.001\} \\
        \bottomrule
    \end{tabular}
\end{table}

\begin{table}[h]
    \centering
    \caption{Optimal Relaxation Hyperparameters for Point Cloud Classification (PE Method).}
    \label{tab:pe_optimal_params}
    \begin{tabular}{llccc}
        \toprule
        \textbf{Model} & \textbf{PE Variant} & \textbf{PE Type} & \textbf{Scheduler} & \textbf{Regularization $\beta$} \\
        \midrule
        VN-PointNet & Additive & Sinusoidal & Linear & 0.001 \\
        VN-PointNet & Concat & Sinusoidal & Linear & 0.005 \\
        \midrule
        VN-DGCNN & Additive & Fixed & Linear & 0.001 \\
        VN-DGCNN & Concat & Sinusoidal & Triangular & 0.001 \\
        \bottomrule
    \end{tabular}
\end{table}
\FloatBarrier

\subsection{Rapidash}
\label{sec:app-rapidash}

We evaluate our PE relaxation method on Rapidash \cite{vadgama_probing_2025}, a regular group-convolution architecture for point clouds defined over the position-orientation space \( \mathbb{R}^3 \times S^2 \).
Following \citet{vadgama_probing_2025}, we use the ModelNet40 classification task \cite{wu_3d_2015} and the strongest strictly equivariant Rapidash configuration considered in their ModelNet40 experiments.
We compare against two Rapidash baselines: Rapidash-Eq, which uses point-cloud normals as vector features and preserves the equivariant test-time architecture, and Rapidash-Symm.\ Break, which additionally uses a global-frame input and therefore breaks the symmetry at inference.
For a controlled comparison with an existing train-time relaxation method on the same backbone, we additionally implement the constraint relaxation (CR) of \citet{pertigkiozoglou_improving_2025} on top of Rapidash-Eq (Rapidash-CR).
As in the Vector Neurons experiments, the relaxation terms are annealed to zero during training, so Rapidash-CR also recovers the strictly equivariant Rapidash-Eq architecture at inference.

\paragraph{Implementation Details.}
We use the original Rapidash ModelNet40 training regieme and hyperparameters from \citet{vadgama_probing_2025}. 
Training uses the original, $z$-aligned ModelNet40 with 1024 points, and each training sample is randomly rotated inside the model pipeline; only Rapidash-Symm.~Break receives this rotation as a global-frame pose feature.
All models use \(C=256\) hidden features and are trained for 120 epochs. 
The Rapidash backbone uses 7 layers, polynomial degree 2, 0 fiber dimensions for the \( \mathbb{R}^3 \) base, and 8 fiber dimensions for the \( \mathbb{R}^3 \times S^2 \) base used in the equivariant ModelNet40 variants. 
Optimization is performed with Adam, learning rate \(10^{-4}\), weight decay \(10^{-8}\), and a cosine-annealing learning-rate schedule with 20 warmup epochs. 
All models are trained on a single NVIDIA RTX A6000 GPU.
The specific hyperparameters are listed in \Cref{tab:rapidash_params}.

\paragraph{Evaluation Protocol.}
We report two validation metrics. 
First, we evaluate on the original ModelNet40 validation set, which reflects the standard orientation-biased benchmark. 
Second, we evaluate on a randomly rotated validation set, where each input is transformed by a random rotation. 
This second metric tests whether the trained model retains the desired rotation robustness rather than exploiting a canonical-orientation bias. 
For Rapidash-Eq + PE, the positional signal is used only during training and is annealed to zero, so the final test-time model recovers the original strictly equivariant Rapidash-Eq architecture.
The same holds for Rapidash-CR.

\paragraph{Hyperparameter Tuning.}
We tune the relaxation-specific hyperparameters for Rapidash-Eq + PE using the same search space as in the Vector Neurons experiments in \Cref{tab:pointnet_sweep}. 
Specifically, we sweep over positional-encoding type, mix-coefficient scheduler, and PE-magnitude regularization strength \(\beta\). 
The search space is shown in \Cref{tab:rapidash_sweep}. 
For Rapidash-CR, we sweep the mix-coefficient scheduler and the regularization strength \(\beta\) over the same values, while keeping all backbone hyperparameters fixed.
The selected configurations are reported in \Cref{tab:rapidash_optimal_params}.
The Rapidash experiments required approximately 92 GPU-hours in total.

\begin{table}[h]
    \centering
    \caption{General Hyperparameters for Rapidash on ModelNet40.}
    \label{tab:rapidash_params}
    \begin{tabular}{lc}
        \toprule
        \textbf{Hyperparameter} & \textbf{Value} \\
        \midrule
        Dataset & ModelNet40 \\
        Input Points & 1024 \\
        Task & Object classification \\
        Hidden Features \(C\) & 256 \\
        Max Epochs & 120 \\
        Rapidash Layers & 7 \\
        Base Space & \( \mathbb{R}^3 \times S^2 \) \\
        \( \mathbb{R}^3 \) Fiber Dimensions & 0 \\
        \( \mathbb{R}^3 \times S^2 \) Fiber Dimensions & 8 \\
        Polynomial Degree & 2 \\
        Optimizer & AdamW \\
        Learning Rate & \(10^{-4}\) \\
        Weight Decay & \(10^{-8}\) \\
        Scheduler & Cosine Annealing \\
        Warmup Epochs & 20 \\
        \bottomrule
    \end{tabular}
\end{table}

\begin{table}[h]
    \centering
    \caption{Hyperparameter Search Space for Rapidash Relaxation (PE and CR).}
    \label{tab:rapidash_sweep}
    \begin{tabular}{lcc}
        \toprule
        \textbf{Hyperparameter} & \textbf{Method} & \textbf{Values Tested} \\
        \midrule
        Positional Encoding Type & PE (Ours) & \{Sinusoidal, Fixed\} \\
        Mix-coef Scheduler & PE / CR & \{Linear, Triangular\} \\
        Regularization \(\beta\) & PE / CR & \{0.1, 0.05, 0.01, 0.005, 0.001\} \\
        \bottomrule
    \end{tabular}
\end{table}

\begin{table}[h]
    \centering
    \caption{Optimal Relaxation Hyperparameters for Rapidash (PE and CR).}
    \label{tab:rapidash_optimal_params}
    \begin{tabular}{lcccc}
        \toprule
        \textbf{Method} & \textbf{PE Variant} & \textbf{PE Type} & \textbf{Scheduler} & \textbf{Regularization \(\beta\)} \\
        \midrule
        Rapidash-Eq+PE & Additive & Sinusoidal & Linear & 0.05 \\
        Rapidash-CR & -- & -- & Triangular & 0.01 \\
        \bottomrule
    \end{tabular}
\end{table}

\FloatBarrier

\subsection{Fractional Shift-Invariant Vision Transformers}
\label{sec:app-xcit}

We evaluate our method on the AF-XCiT-Nano architecture \cite{michaeli_alias-free_2025} trained on the ImageNet dataset \cite{deng_imagenet_2009}.
This architecture is designed to be invariant to fractional cyclic translations, a property we relax during training by reintroducing standard Fourier Positional Encodings used in the XCiT architecture \cite{el-nouby_xcit_2021}.

\paragraph{Implementation Details.}
All models are trained on 8 NVIDIA RTX A6000 GPUs.
We use the original training hyperparameters described by \citet{michaeli_alias-free_2025}. 
We utilize a linear relaxation decay schedule for the first 320 epochs, followed by 80 epochs where the positional encoding coefficient $\lambda$ is set to 0 to recover strict invariance. 
The regularization term is set to $\beta = 10^{-5}$. The full set of hyperparameters is listed in \Cref{tab:xcit_params}.

\paragraph{Hyperparameter Tuning.}
The relaxation hyperparameters (decay scheduler and regularization) were selected after tuning on the smaller Stanford Cars dataset \cite{krause_3d_2013}. 
The search space is detailed in \Cref{tab:xcit_sweep}. 
Based on this sweep, we chose the linear decay schedule and $\beta = 10^{-5}$ for the final ImageNet training runs.
The AF-XCiT experiments, including full training on ImageNet, required approximately 6,700 GPU-hours in total.

\begin{table}[h]
    \centering
    \caption{General Hyperparameters for AF-XCiT (ImageNet).}
    \label{tab:xcit_params}
    \begin{tabular}{lc}
        \toprule
        \textbf{Hyperparameter} & \textbf{Value} \\
        \midrule
        Architecture & AF-XCiT-Nano \\
        Total Epochs & 400 \\
        Relaxation Decay Epochs & 320 \\
        Strict Equivariance Epochs & 80 \\
        Regularization $\beta$ & $10^{-5}$ \\
        Batch Size & 512 \\
        Optimizer & AdamW \\
        Learning Rate & $5 \cdot 10^{-4}$ \\
        \bottomrule
    \end{tabular}
\end{table}

\begin{table}[h]
    \centering
    \caption{Hyperparameter Search Space for AF-XCiT (Tuned on Stanford Cars).}
    \label{tab:xcit_sweep}
    \begin{tabular}{lc}
        \toprule
        \textbf{Hyperparameter} & \textbf{Values Tested} \\
        \midrule
        Regularization $\beta$ & $\{0.1, 0.01, 10^{-3}, 10^{-4}, 10^{-5}\}$ \\
        Decay Schedule & \{Linear, Triangular\} \\
        \bottomrule
    \end{tabular}
\end{table}
\FloatBarrier

\subsection{Smoke Plume Forecasting (E2CNN)}
\label{sec:app-smoke}

We evaluate our method on the synthetic smoke plume dataset with approximate rotation symmetry of \citet{wang_approximately_2022}.
The dataset consists of $64 \times 64$ 2D smoke velocity fields simulated with PhiFlow, where the inflow locations and buoyant force directions are $C_4$-symmetric, while the buoyancy factor varies with the inflow position and breaks this symmetry.
Following the official implementation of \citet{wang_approximately_2022}, we use 25 simulations for training and 10 held-out simulations for the Domain test set.
We compare against two baselines: the strictly $C_4$-equivariant E2CNN \cite{weiler_general_2019}, and RSteer \cite{wang_approximately_2022}, which relaxes the steerable-kernel constraint and remains approximately equivariant at inference.
We evaluate two variants of our method on top of E2CNN: decayed PE, which anneals the PE to zero and recovers the strict E2CNN at inference, and persistent PE, which keeps the PE at inference.

\paragraph{Implementation Details.}
We build on the official codebase of \citet{wang_approximately_2022} and use its Rotation training hyperparameters for all models.
The E2CNN backbone uses 5 layers with $3 \times 3$ kernels and 92 regular-representation channels of $C_4$ (approximately 0.62M parameters, compared to 5.6M for RSteer).
Models receive 10 input frames, are trained with a loss over 6 autoregressive steps, and are optimized with Adam, learning rate $10^{-3}$, weight decay $4 \cdot 10^{-4}$, and a step learning-rate schedule with decay factor $0.95$ per epoch.
Training runs for up to 1000 epochs with the official early-stopping rule, which is applied from epoch 100 onward; in practice, all runs stopped between epochs 101 and 107.
For RSteer, we use the official weight-relaxation coefficient $\alpha = 10^{-5}$.
For our method, we add the 2D Fourier PE described in \Cref{sec:experiments} to the input velocity field.
All models are trained on a single NVIDIA RTX PRO 6000 Blackwell GPU.
The specific hyperparameters are listed in \Cref{tab:smoke_params}.

\paragraph{Evaluation Protocol.}
We report the 20-step autoregressive prediction RMSE on two test sets: Future, which contains later time steps (samples 200--250) of the training simulations, and Domain, which contains the held-out simulations.
Samples 160--200 of the training simulations are used for validation, and samples 0--160 for training.
For all models, we evaluate the checkpoint with the lowest validation RMSE.
For decayed PE, the PE is removed at test time, so the evaluated model is exactly the strictly equivariant E2CNN; for persistent PE, the PE is kept at test time.
We report mean $\pm$ std over 3 random seeds.

\paragraph{Hyperparameter Tuning.}
We tune the relaxation-specific hyperparameters of decayed PE on a single seed, selecting the configuration with the lowest validation RMSE with PE removed.
We sweep over the mix-coefficient scheduler, the schedule length, and the PE-magnitude regularization strength \(\beta\), using the same \(\beta\) values as in the Vector Neurons experiments.
Both schedule lengths reach \(\lambda = 0\) before early stopping can be applied.
The search space is shown in \Cref{tab:smoke_sweep}, and the selected configuration is reported in \Cref{tab:smoke_optimal_params}.
The persistent-PE variant uses no decay and no regularization.
The backbone hyperparameters are not tuned for any of the models.
The smoke plume experiments required approximately 56 GPU-hours in total.

\begin{table}[h]
    \centering
    \caption{General Hyperparameters for Smoke Plume Forecasting (E2CNN).}
    \label{tab:smoke_params}
    \begin{tabular}{lc}
        \toprule
        \textbf{Hyperparameter} & \textbf{Value} \\
        \midrule
        Dataset & Smoke plumes (Rotation) \\
        Task & Velocity-field forecasting \\
        Symmetry Group & \(C_4\) \\
        Layers & 5 \\
        Hidden Channels & 92 \\
        Kernel Size & \(3 \times 3\) \\
        Input Frames & 10 \\
        Training Rollout Steps & 6 \\
        Test Rollout Steps & 20 \\
        Batch Size & 16 \\
        Max Epochs & 1000 (early stopping) \\
        Optimizer & Adam \\
        Learning Rate & \(10^{-3}\) \\
        Weight Decay & \(4 \cdot 10^{-4}\) \\
        Scheduler & Step decay (\(\times 0.95\) per epoch) \\
        RSteer \(\alpha\) & \(10^{-5}\) \\
        \bottomrule
    \end{tabular}
\end{table}

\begin{table}[h]
    \centering
    \caption{Hyperparameter Search Space for Smoke Plume PE Relaxation.}
    \label{tab:smoke_sweep}
    \begin{tabular}{lc}
        \toprule
        \textbf{Hyperparameter} & \textbf{Values Tested} \\
        \midrule
        Mix-coef Scheduler & \{Linear, Triangular\} \\
        Schedule Length (epochs) & \{50, 100\} \\
        Regularization \(\beta\) & \{0.1, 0.05, 0.01, 0.005, 0.001\} \\
        \bottomrule
    \end{tabular}
\end{table}

\begin{table}[h]
    \centering
    \caption{Optimal Relaxation Hyperparameters for Smoke Plume PE Relaxation.}
    \label{tab:smoke_optimal_params}
    \begin{tabular}{lccc}
        \toprule
        \textbf{Variant} & \textbf{Scheduler} & \textbf{Schedule Length} & \textbf{Regularization \(\beta\)} \\
        \midrule
        Decayed PE & Linear & 50 & 0.005 \\
        Persistent PE & None & -- & 0 \\
        \bottomrule
    \end{tabular}
\end{table}
\FloatBarrier

\subsection{N-body Simulation (SEGNN)}
\label{sec:app-segnn}

We evaluate our approach on the charged-particle N-Body forecasting task \cite{kipf_neural_2018} using the SEGNN architecture \cite{brandstetter_geometric_2022}. 
The task involves predicting the future positions of 5 charged particles in a 3D space.

\paragraph{Experimental Setup.}
Following  \citet{pertigkiozoglou_improving_2025}, we use a dataset split containing 3000 training samples, 2000 validation samples, and 2000 test samples. 
We train for 2000 epochs and report the test MSE on final particle positions for the checkpoint with the lowest validation loss.
All models are trained on a single NVIDIA RTX A6000 GPU.

\paragraph{Hyperparameter Tuning.}
We perform a two-stage hyperparameter search. 
First, we tune the optimization hyperparameters (learning rate and weight decay) for the strictly equivariant SEGNN baseline to establish a strong baseline. 
Based on this sweep (\Cref{tab:segnn_optim_sweep}), we fix the learning rate to $2 \cdot 10^{-3}$ and weight decay to $10^{-8}$ for all subsequent relaxation experiments.

Next, we tune the relaxation-specific parameters for both the baseline Constraint Relaxation (CR) method and our proposed PE method using the fixed optimization parameters. 
For our method, we additionally sweep over the positional encoding type. 
The search spaces are detailed in \Cref{tab:segnn_relax_sweep}.

The optimal configurations resulting from this tuning stage are listed in \Cref{tab:segnn_optimal_params}.
For the Constraint Relaxation (CR) baseline, we found that a linear schedule with $\beta=0.1$ performed best.
For our PE method, a triangular schedule with $\beta=0.05$ and learned (fixed) positional encoding yielded the optimal results.
The SEGNN experiments required approximately 13 GPU-hours in total.

\begin{table}[h]
    \centering
    \caption{General Hyperparameters for SEGNN (N-Body).}
    \label{tab:segnn_params}
    \begin{tabular}{lc}
        \toprule
        \textbf{Hyperparameter} & \textbf{Value} \\
        \midrule
        Dataset & Charged-particle N-body \\
        Split (Train/Val/Test) & 3000 / 2000 / 2000 \\
        Particles & 5 \\
        Batch Size & 100 \\
        Max Epochs & 2000 \\
        Optimizer & Adam \\
        Learning Rate (Final) & $2 \cdot 10^{-3}$ \\
        Weight Decay (Final) & $10^{-8}$ \\
        \bottomrule
    \end{tabular}
\end{table}

\begin{table}[h]
    \centering
    \caption{Optimization Hyperparameter Search Space (SEGNN Strict Baseline).}
    \label{tab:segnn_optim_sweep}
    \begin{tabular}{lc}
        \toprule
        \textbf{Hyperparameter} & \textbf{Values Tested} \\
        \midrule
        Learning Rate & $\{5 \cdot 10^{-4}, 8 \cdot 10^{-4}, 1 \cdot 10^{-3}, 2 \cdot 10^{-3}, 5 \cdot 10^{-3}\}$ \\
        Weight Decay & $\{10^{-12}, 10^{-8}, 10^{-6}, 10^{-4}\}$ \\
        \bottomrule
    \end{tabular}
\end{table}

\begin{table}[h]
    \centering
    \caption{Relaxation Hyperparameter Search Space (SEGNN).}
    \label{tab:segnn_relax_sweep}
    \begin{tabular}{lcc}
        \toprule
        \textbf{Hyperparameter} & \textbf{Method} & \textbf{Values Tested} \\
        \midrule
        Mix-coef Scheduler & PE / CR & \{Linear, Triangular\} \\
        Regularization $\beta$ & PE / CR & \{0.1, 0.05, 0.01, 0.005, 0.001\} \\
        PE Type & PE (Ours) & \{Sinusoidal, Learned\} \\
        \bottomrule
    \end{tabular}
\end{table}

\begin{table}[h]
    \centering
    \caption{Optimal Relaxation Hyperparameters for SEGNN.}
    \label{tab:segnn_optimal_params}
    \begin{tabular}{lccc}
        \toprule
        \textbf{Method} & \textbf{Scheduler} & \textbf{Regularization $\beta$} & \textbf{PE Type} \\
        \midrule
        CR & Linear & 0.1 & - \\
        PE (Ours) & Triangular & 0.05 & Learned \\
        \bottomrule
    \end{tabular}
\end{table}

\FloatBarrier

\section{Dataset Assets, Licenses, and Terms of Use}
\label{app:dataset-licenses}

All datasets used in this work are public and widely used benchmarks in academic machine learning research. We cite the original dataset papers in the main text and summarize the known license or terms-of-use information below, to the best of our knowledge. We use these datasets only for academic research and evaluation, and we do not redistribute the original dataset contents in our supplementary material.

\begin{itemize}
        \item \textbf{ModelNet40} \citep{wu_3d_2015}. 
    We use ModelNet40 for point-cloud classification experiments with Vector Neurons and Rapidash. In our experiments, ModelNet40 is accessed through PyTorch Geometric. The PyTorch Geometric loader code is released under the MIT license. The loader downloads the original ModelNet40 CAD dataset from the official Princeton ModelNet URL. The ModelNet website states that all CAD models were downloaded from the Internet, that the original authors hold the copyright of the CAD models, that the labels were obtained via Amazon Mechanical Turk and are provided freely, and that the dataset is provided for the convenience of academic research only. We use ModelNet40 only for academic evaluation and do not redistribute the raw CAD models.

    \item \textbf{ImageNet / ImageNet-1k} \citep{deng_imagenet_2009}. 
    We use ImageNet for image classification experiments with AF-XCiT and XCiT. ImageNet is publicly accessible under the ImageNet access agreement. The ImageNet terms specify non-commercial research and educational use, no warranty, user responsibility for use of copyrighted images, and access sharing only with collaborators who agree to the same terms. We use ImageNet only for non-commercial academic benchmarking and do not redistribute the images.

        \item \textbf{Smoke plume (rotation)} \citep{wang_approximately_2022}.
    We use the synthetic smoke plume dataset with approximate rotation symmetry for the E2CNN. 
    The dataset is simulated with the PhiFlow fluid solver, using the data-generation notebook released with the official code of \citet{wang_approximately_2022}, which also provides a preprocessed copy. To the best of our knowledge, the official repository does not include an explicit license file; PhiFlow is released under the MIT license. The dataset is synthetic and contains no human-subject or sensitive data. We use it only for academic evaluation and do not redistribute it.

    \item \textbf{Stanford Cars / Cars196} \citep{krause_3d_2013}. 
    We use Stanford Cars for hyperparameter tuning of the AF-XCiT relaxation schedule and regularization before the final ImageNet runs. Stanford Cars is a public and widely used academic fine-grained recognition benchmark. To the best of our knowledge, we did not find an explicit dataset-specific license in the current public metadata; standard dataset loaders and catalog entries provide the citation and dataset description, while the original hosting URL is no longer reliably available. We use the dataset only for academic hyperparameter tuning and do not redistribute the images.

    \item \textbf{Charged-particle N-body benchmark} \citep{kipf_neural_2018}. 
    We use the charged-particle N-body benchmark for SEGNN forecasting experiments. This is a synthetic benchmark generated following the Neural Relational Inference (NRI) setup. The official NRI code release provides scripts to generate charged-particle simulations and is released under the MIT license. We generate and use the synthetic benchmark for academic evaluation and do not introduce or redistribute any human-subject or sensitive data.
\end{itemize}

\end{document}